\pdfoutput=1
\documentclass[10pt,twocolumn,letterpaper]{article}

\usepackage{arxiv}

\usepackage[normalem]{ulem}
\usepackage{nicefrac}
\usepackage{microtype}

\usepackage{cuted}

\usepackage{array}
\usepackage{multirow}
\usepackage{makecell}
\usepackage{threeparttable}
\usepackage{siunitx}
\usepackage{tabularx}
\usepackage{adjustbox}

\usepackage{colortbl}
\usepackage{arydshln}

\usepackage[most]{tcolorbox}
\usepackage{varwidth}

\usepackage{algorithm}
\usepackage{algpseudocode}

\algrenewcommand\algorithmicrequire{\textbf{Input:}}
\algrenewcommand\algorithmicensure{\textbf{Output:}}

\definecolor{sectiongray}{RGB}{238,238,238}

\definecolor{PWExplicitBG}{RGB}{234,244,252}
\definecolor{PWImplicitBG}{RGB}{236,247,239}
\definecolor{PWSectionBG}{RGB}{246,247,249}

\definecolor{colorFst}{HTML}{F59194}
\definecolor{colorSnd}{HTML}{FAC791}
\definecolor{colorThd}{HTML}{FFFF99}
\definecolor{url_color}{RGB}{42,83,163}

\definecolor{PromptPurple}{RGB}{110,40,130}
\definecolor{PromptBorder}{RGB}{140,90,160}
\definecolor{PromptBg}{RGB}{252,248,255}

\newtcolorbox{promptbox}[2][]{%
    enhanced,
    colback=PromptBg,
    colframe=PromptBorder,
    boxrule=0.8pt,
    arc=2mm,
    left=1.2mm,
    right=1.2mm,
    top=1.2mm,
    bottom=1.2mm,
    fonttitle=\bfseries\small,
    coltitle=white,
    colbacktitle=PromptPurple,
    title={#2},
    attach boxed title to top left={
        xshift=2mm,
        yshift*=-2mm
    },
    boxed title style={
        sharp corners,
        arc=1mm,
        boxrule=0pt,
        colback=PromptPurple
    },
    before skip=6pt,
    after skip=6pt,
    #1
}

\AtBeginDocument{%
}

\definecolor{linkblue}{rgb}{0.21,0.49,0.74}
\usepackage[pagebackref,breaklinks,colorlinks,allcolors=linkblue]{hyperref}

\title{MorphTwin: Enriching Deformable Material Responses from Videos for Morphable and Interactive Physical Twins}

\author{%
Haoqin Hong$^{1,2*}$\quad
Ding Fan$^{1*}$\quad
Fubin Dou$^{1*}$\quad
Tianyu Huang$^{2\ddagger}$\quad
Yuzhuo Chen$^{1}$\quad
Zhemeng Zhang$^{3}$\\
Hanxiao Jiang$^{4}$\quad
Congcong Zhu$^{1\dagger}$\quad
Gerhard Petrus Hancke$^{2}$\quad
Jingrun Chen$^{1\dagger}$\quad
Rynson W.H. Lau$^{2}$\\[0.6em]
$^{1}$University of Science and Technology of China\qquad
$^{2}$City University of Hong Kong\\
$^{3}$Multimedia Laboratory, The University of Hong Kong\qquad
$^{4}$Columbia University\\[0.4em]
{\small $^{*}$Equal contribution\qquad
$^{\dagger}$Corresponding authors\qquad
$^{\ddagger}$Project leader}
}

\begin{document}
\maketitle

\begin{strip}
    \centering
    \includegraphics[width=\linewidth]{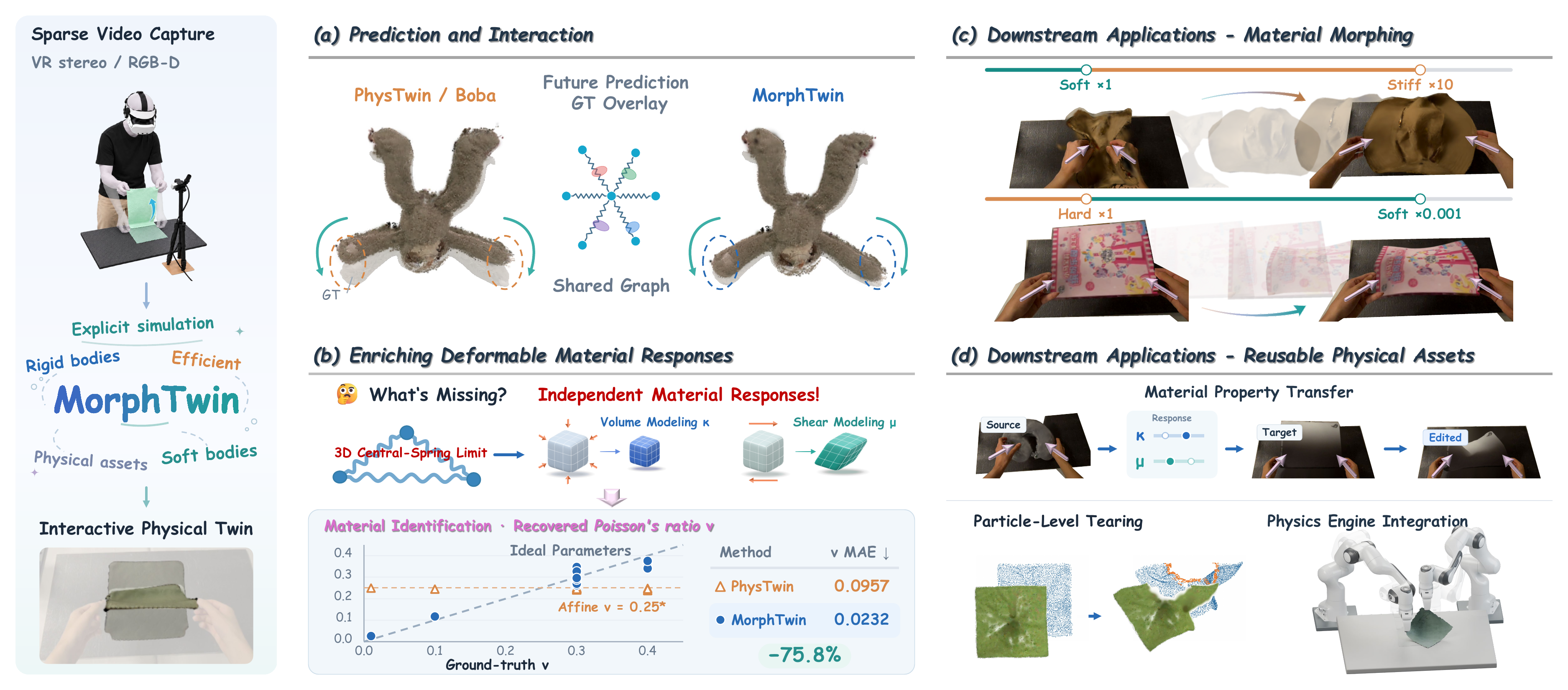}

    \captionsetup{hypcap=false}
    \captionof{figure}{
        \textbf{MorphTwin expands the expressive capacity of spring-mass models for deformable material responses while preserving their efficient explicit simulation.}
        Sparse videos yield interactive physical twins.
        \textbf{\textit{(a) Prediction and Interaction.}}
        MorphTwin more faithfully predicts future soft-body deformations than PhysTwin~\citep{jiang2025phystwin} and Boba~\citep{pang2026boba} while retaining fast explicit dynamics for real-time interaction.
        \textbf{\textit{(b) Enriching Deformable Material Responses.}}
        Decoupled volumetric and shear modeling relaxes the isotropic affine constraint ($\nu=0.25$) of 3D central-force springs, improving Poisson's ratio recovery.
        \textbf{\textit{(c) Material Morphing.}}
        Editing material responses stiffens a fabric hat and softens a book while preserving their initial geometry and appearance.
        \textbf{\textit{(d) Reusable Physical Assets.}}
        Transferring material responses from a soft source softens a hard mousepad.
        Reconstructed assets support particle-level tearing and physics-engine integration.
    }
    \label{fig:teaser}
    \captionsetup{hypcap=true}
\end{strip}

% \begin{abstract}
% Reconstructing interactive physical twins from video requires balancing material response accuracy with simulation efficiency. While mass-spring systems support fast explicit simulation, the volumetric and shape responses of classical central-force springs can be coupled, and their spring parameters depend on graph geometry and connectivity, limiting material recovery and transfer. We present MorphTwin, which enriches local deformation modes to enable independent control of volumetric and shape responses while retaining simple spring forces and explicit forward simulation. Furthermore, we represent material adjustments as shared relative changes in material responses and map them to graph-specific stiffness updates via offline calibration, supporting interpretable material editing and cross-object transfer. Under identical inverse settings on synthetic scenes, MorphTwin reduces the mean absolute error of Poisson's ratio by 75.8\% and the mean relative error of Young's modulus by 22.1\% compared to PhysTwin. We introduce a benchmark comprising synchronized egocentric stereo and RGB-D interaction sequences. Across diverse real-world experiments, MorphTwin improves reconstruction and future motion prediction while retaining real-time simulation.
% \end{abstract}

\begin{abstract}
Reconstructing interactive physical twins from video requires
accurately capturing how objects deform while maintaining
efficient simulation.
We observe that reproducing motion in a video does not guarantee
correct responses to new forces when the physical representation
restricts how an object can deform.
In spring-based twins, coupling between volume and shape
responses creates such a limitation.
We present \textbf{MorphTwin}, which enriches local deformation
modes to capture and independently control these responses
while retaining standard spring forces and explicit simulation.
To support interpretable material editing and transfer,
we use offline calibration to map shared relative response
changes to stiffness updates on each reconstructed graph.
Under identical inverse settings on synthetic scenes,
MorphTwin reduces the mean absolute error of Poisson's ratio
by 75.8\% and the mean relative error of Young's modulus
by 22.1\% compared to PhysTwin.
We introduce an egocentric stereo and RGB-D benchmark covering
interactions with objects of varying stiffness.
Real-world experiments demonstrate improved reconstruction
accuracy and up to 14.4\% lower future tracking error
than PhysTwin at interactive speeds.
\end{abstract}
\section{Introduction}
\label{sec:introduction}

Reconstructing physical twins from videos turns observed objects into interactive assets that respond to novel forces and manipulations. Applications such as virtual reality and embodied AI require physically plausible responses at interactive speeds~\citep{ai2025robot_review_science_robotics,spielberg2023inverse_problem_reviews_nature_machine_intelligence,frisoli2024vr_review_nature_reviews}. Continuum methods such as PhysGaussian~\citep{xie2024physgaussian} and GIC~\citep{cai2024gic} use Material Point Methods (MPM)~\citep{jiang2016mpm} to model rich material behaviors, but detailed simulations can be costly for real-time interaction. PGND~\citep{zhang2025pgnd} learns particle motion but offers limited explicit control over material properties, while DeformMaster~\citep{li2026deformmaster} combines MPM simulator with neural corrections and faces a trade-off between simulation accuracy and interactive speed. Mass-spring systems~\citep{blickhan1989mss} provide a real-time foundation for explicit forward simulation, with simple local forces and directly adjustable mechanical parameters. PhysTwin~\citep{jiang2025phystwin,zhong2024springgaussian} combines this structure
with visual reconstruction, enabling physical twin reconstruction
from sparse-view videos and real-time interaction, but the
underlying spring representation can limit material expressiveness.

Our goal is to turn spring-based physical twins into physically
faithful, reusable assets with interpretable material control.
We further observe that these models can closely match
observed motion yet respond incorrectly to novel interactions.
This motivates an elastic-energy analysis of conventional
central-force networks~\citep{golec2020hybridMSS}.
As illustrated in Fig.~\ref{fig:teaser}(b), we identify a source
of this discrepancy: their responses to volume and shape
changes can be coupled.
Consequently, a reconstructed model may capture
how far an object stretches yet fail to reproduce how much it thins.
Both responses govern shape and volume changes in deformable objects.
This structural coupling can persist despite stiffness fitting.
Overcoming it requires accommodating irregular geometry
reconstructed from sparse views while retaining standard
spring forces and efficient explicit simulation.

% We observe that spring-based physical twins can closely match
% observed motion yet respond incorrectly to novel interactions.
% This motivates an elastic-energy analysis of conventional
% central-force networks~\citep{golec2020hybridMSS}.
% As illustrated in Fig.~\ref{fig:teaser}(b), we identify a source
% of this discrepancy: their responses to volume and shape
% changes can be coupled.
% Consequently, a reconstructed model may capture
% how far an object stretches yet fail to reproduce how much it thins.
% Both responses govern shape and volume changes in deformable objects.
% This structural coupling can persist despite stiffness fitting.
% Overcoming it requires accommodating irregular geometry
% reconstructed from sparse views while retaining standard
% spring forces and efficient explicit simulation.

Accordingly, we introduce MorphTwin, a framework that enriches
the deformation modes of reconstructed spring graphs to support
independent estimation and control of volume and shape responses.
We augment each graph with local structures whose internal
nodes can move relative to surrounding nodes.
These response carriers introduce additional deformation
freedom while retaining standard central-force springs.
All nodes evolve together through explicit forward simulation,
preserving efficient interaction.

To support interpretable editing and transfer of material
behavior, we account for how graph geometry and
connectivity affect mechanical responses.
We calibrate each graph using standardized volumetric and shear
deformation tests, mapping shared relative response changes
to graph-specific stiffness updates.
This enables consistent material adjustments across objects
relative to their respective reference states.
Calibration is performed offline, and online interaction
proceeds through explicit spring simulation.

Controlled experiments show that our method improves material recovery accuracy and better preserves relative material responses across different graph discretizations. Real-world experiments further demonstrate that MorphTwin improves both reconstruction and future prediction while maintaining efficient explicit simulation, and supports reconstructing physical twins from egocentric observations. Together, these results enable interactive physical twins with richer, editable material behaviors.

% Our main contributions are as follows:
% \begin{itemize}
%     \item \textbf{1)~Expressing richer material responses in explicit spring dynamics.} We analyze the coupling between bulk and shear responses in classical central-force networks and design response carriers with internal degrees of freedom to enable independent control over both.
%     \item \textbf{2)~Response calibration for material identification and transfer.} We calibrate each graph using elastic energy under standardized deformations, mapping shared relative material response changes to graph-specific spring parameters to support material identification, editing, and transfer across objects and discretizations.
%     \item \textbf{3)~A framework for morphable and interactive physical twins.}
%     MorphTwin integrates material identification, editing, and reuse
%     from videos while retaining efficient explicit simulation.
%     On synthetic scenes with known parameters, MorphTwin identifies
%     physically meaningful Poisson ratios and estimates Young's
%     modulus more accurately.
%     We also introduce a benchmark combining egocentric stereo and RGB-D
%     observations, and validate improved reconstruction and future
%     prediction on PhysTwin and our benchmark, including single
%     and sparse view settings.
% \end{itemize}

Our main contributions are as follows:
\begin{itemize}
    \item \textbf{Reconstructing physical twins with richer material responses.}
    We present MorphTwin, which enriches local deformation modes
    during video-based reconstruction to enable independent
    volume and shape control while retaining efficient
    explicit spring dynamics.

    \item \textbf{Response carriers and calibration for material control.}
    We design internally relaxable spring structures and
    energy-based calibration to support material identification
    and map shared relative response changes to graph-specific
    stiffness updates for interpretable editing and transfer.

    \item \textbf{Controlled validation and an egocentric benchmark.}
    We demonstrate improved material recovery, reconstruction,
    and future prediction through controlled experiments
    and real-world evaluations.
    We introduce a benchmark of temporally aligned egocentric
    stereo and RGB-D interaction sequences and demonstrate
    material editing and transfer in interactive physical twins.
\end{itemize}
\section{Related Work}
\label{sec:related_work}

\textbf{Video-based physical twins and interactive dynamics.}
Recent methods reconstruct deformable physical twins from visual
observations~\citep{li2023pac,cai2024gic,jiang2025phystwin,rho2026monophysics,kim2026egophys,lee2026physhandi,hong2026pidg}.
PhysTwin~\citep{jiang2025phystwin} combines Gaussian appearance with
spring-mass dynamics. PGND~\citep{zhang2025pgnd} and
DeformMaster~\citep{li2026deformmaster} improve prediction through
learned dynamics and residual corrections, while
Boba~\citep{pang2026boba} accelerates simulation with compact models
and batched execution.
Yet accurate motion fitting and efficient rollout do not ensure
physically meaningful, reusable material parameters.
MorphTwin identifies interpretable material responses while keeping
material calibration offline and online simulation explicit
for efficient interactive rollout.

\textbf{Material representations and reuse across graphs.}
Spring parameters depend on graph geometry and
connectivity~\citep{lloyd2008identification}.
MatPhys~\citep{yang2026matphys} learns shared material embeddings,
while NeuSpring~\citep{xu2026neuspring} models spring properties
with neural fields. However, consistent parameter prediction does
not guarantee consistent mechanical responses across graphs.
Central springs also couple bulk and shear in unstressed,
isotropic 3D networks under affine small strains.
MorphTwin uses internally relaxable response carriers for separate
bulk and shear control within standard central-force dynamics.
Energy alignment maps shared response targets to graph-specific
spring parameters for interpretable recovery, editing, and reuse
across objects and discretizations.
Appendix~\ref{app:clarify} compares the inputs,
processing, and outputs of related methods, highlighting
MorphTwin's advantages.

\section{Problem Formulation}
\label{sec:problem_formulation}

\paragraph{Problem setting.}
Given an interaction video $\mathcal V$ and its reconstructed
spring-mass graph $\mathcal G=(\mathcal P,\mathcal S)$, where
$\mathcal P$ and $\mathcal S$ denote physical nodes and
central-force spring connections, our goal is to identify
physically meaningful material responses from limited
deformation observations.
The resulting twins should generalize to novel forces and
manipulations and support material editing and reuse at
interactive speeds.
While FEM~\citep{zienkiewicz2013fem} and
MPM~\citep{jiang2016mpm} offer expressive constitutive models,
their costs can grow substantially at fine resolutions.
We build on the spring-mass architecture of
PhysTwin~\citep{jiang2025phystwin} to retain efficient
explicit dynamics.

\paragraph{Material parameters and graph discretization.}
Spring coefficients are defined on a specific graph, and
their mechanical effect depends on its geometry and connectivity.
Discretization and material identification are therefore coupled:
under comparable geometry and loading, compliant objects
with rich local deformations may require substantially more
simulation nodes and spring connections than stiffer objects.
For an unstressed graph, the affine elastic energy $\mathcal E^{\mathrm{aff}}$ under
a small-strain tensor $\boldsymbol\epsilon$ is
\begin{equation*}
\mathcal E^{\mathrm{aff}}(\boldsymbol\epsilon)
= \frac{1}{2}\sum_{e\in\mathcal S}Y_{e}L_{e}
(\mathbf n_{e}^{\top}\boldsymbol\epsilon\mathbf n_{e})^2,
\end{equation*}
where $Y_e=k_eL_e$, $k_e$ is Hookean spring constant,
and $L_e,\mathbf n_e$ are edge's reference length
and unit direction.
Because this energy also depends on graph geometry and
connectivity, identical coefficients can produce different
mechanical responses across discretizations.
Material identification and reuse therefore require an
explicit correspondence between physical responses and
graph-specific spring parameters.

\paragraph{Expressiveness of central-force networks.}
Material recovery also depends on which responses a graph
can express.
For an unstressed three-dimensional central-force network
with an isotropic effective response under affine small strains,
directional averaging gives
\begin{equation}
\begin{aligned}
&W_{\rm central}
=
\frac{c}{30}
\left[
(\operatorname{tr}\boldsymbol\epsilon)^2
+
2\boldsymbol\epsilon:\boldsymbol\epsilon
\right], \\
&W_{\rm iso}
=
\frac{\lambda}{2}
(\operatorname{tr}\boldsymbol\epsilon)^2
+
\mu\boldsymbol\epsilon:\boldsymbol\epsilon
=
\underbrace{\frac{\lambda}{2}}_{=\,c/30}
\left[
(\operatorname{tr}\boldsymbol\epsilon)^2
+
2\underbrace{\frac{\mu}{\lambda}}_{=\,1}
\boldsymbol\epsilon:\boldsymbol\epsilon
\right].
\end{aligned}
\label{eq:central_vs_iso}
\end{equation}
Here, $c$ is the effective network stiffness scale,
$\lambda,\mu$ are the Lam\'e parameters, $\mu$ is the shear
modulus, and $:$ denotes tensor double contraction.
Matching the coefficients yields
\begin{equation}
\lambda=\mu=\frac{c}{15},
\quad
\kappa=\lambda+\frac{2}{3}\mu=\frac{5}{3}\mu,
\quad
\nu=\frac{3\kappa-2\mu}{2(3\kappa+\mu)}=\frac{1}{4},
\label{eq:cauchy_limit}
\end{equation}
where $\kappa$ is the bulk modulus and $\nu$ is Poisson's ratio.
Topology and stiffness changes within this model class
preserve the fixed ratio $\kappa/\mu=5/3$, preventing
independent control of volume and shape responses.
Consequently, fitting axial stiffness alone cannot recover
the correct transverse response of materials with $\nu\neq1/4$.

These observations motivate MorphTwin's response carriers
and graph-specific energy calibration.
Internally relaxable carriers enrich controllable local deformation modes,
while offline calibration maps desired relative bulk and
shear response changes to graph-specific stiffness updates.
Together, they support material identification, editing,
and reuse across different reconstructed graphs while
retaining standard central-force springs and efficient
explicit integration.
\section{Methodology}
\label{sec:methodology}
\begin{figure*}[t]
\centering
\includegraphics[width=\textwidth]{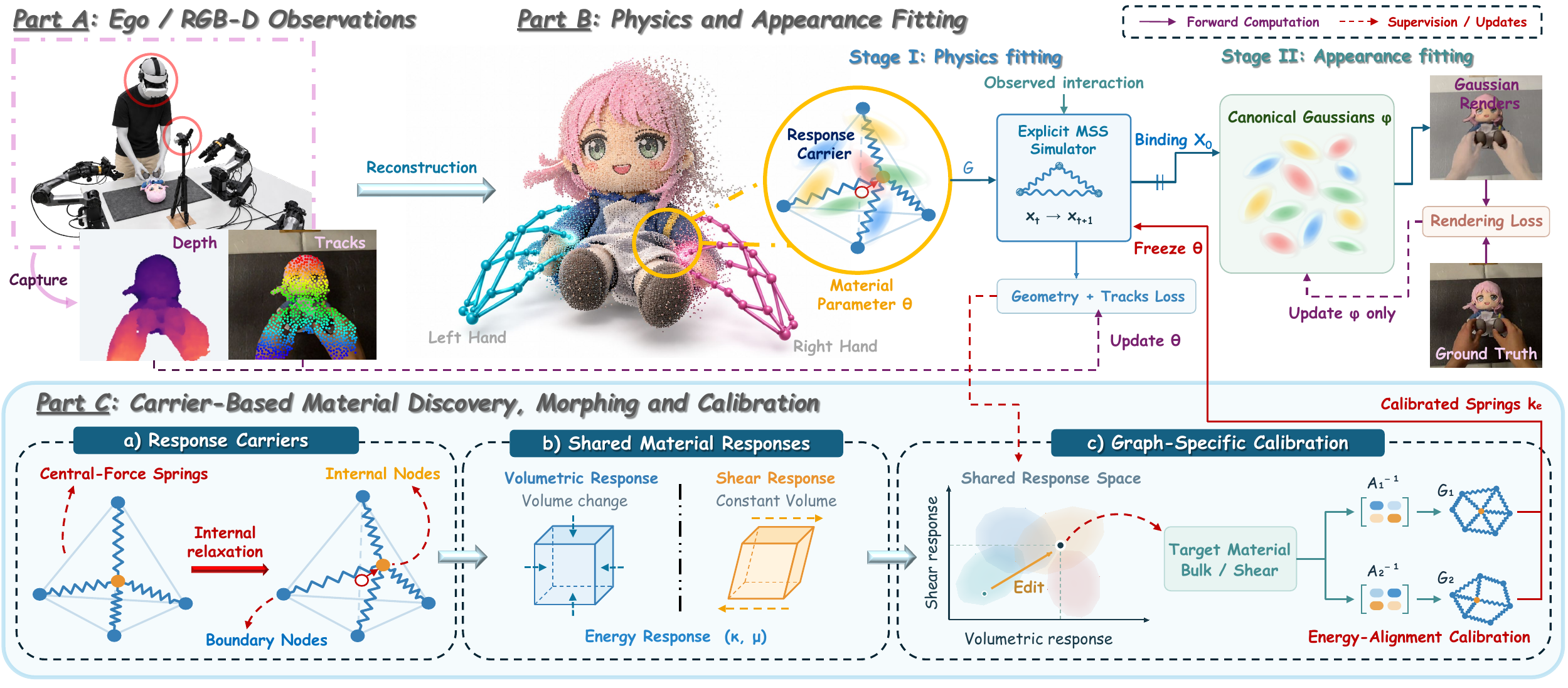}
\caption{\textbf{Overview of MorphTwin.}
\underline{\textit{Part A:}} Sparse video observations
provide geometry and motion cues.
\underline{\textit{Part B:}} We fit the physical model
to observed motion, followed by Gaussian appearance
optimization with physics fixed.
\underline{\textit{Part C:}} Relaxable response carriers
enable separate volumetric and shear control.
Offline calibration maps shared relative response changes
to spring stiffnesses for each graph, enabling material
editing and reuse. Online interaction retains explicit
spring dynamics.}
\label{fig:morphtwin-framework}
\end{figure*}

Figure~\ref{fig:morphtwin-framework} provides an overview of our MorphTwin pipeline.
We first initialize an explicit spring graph from geometry
and motion cues extracted from sparse video observations
(Sec.~\ref{sec:physics_twin}). We then introduce relaxable response carriers and calibrate
the graph under volumetric and shear deformations (Secs.~\ref{sec:material_response}-\ref{sec:material_identification}). The calibrated model is fitted to observed motion, followed
by Gaussian appearance optimization with physics fixed.
Material edits are expressed as shared relative response
changes and mapped to spring stiffnesses for each graph,
enabling material editing and reuse while retaining
efficient explicit simulation.
(Sec.~\ref{sec:editing}).

\subsection{Video and Graph Initialization}
\label{sec:physics_twin}

Given sparse video observations $\mathcal V$, we follow
PhysTwin~\citep{jiang2025phystwin} to obtain object masks
with Grounded-SAM~2~\citep{liu2024grounding,ravi2025sam2}
and 2D point tracks with
CoTracker3~\citep{karaev2025cotracker3}.
We lift these tracks to 3D using the corresponding depth
maps and align a shape prior from
TRELLIS~\citep{xiang2025trellis}
to the initial observations.

Spatial neighbor connections initialize the spring graph
$\mathcal G_0=(\mathcal P_0,\mathcal S_0)$, with physical nodes
$\mathcal P_0$ and springs $\mathcal S_0$.
We denote the backbone stiffnesses by $\mathbf b$. The interaction parameters $\eta$ govern the stiffness of springs
linking the hand to the object, as well as friction and restitution
during contact.
The recovered geometry and 3D tracks provide supervision
for subsequent physics fitting.

% Given spring stiffnesses $\mathbf Y$, node $i$ evolves according to $m_i\ddot{\mathbf x}_i^t=\mathbf f_i^{\mathrm{spr}}(\mathbf X^t;\mathbf Y)+\mathbf f_i^{\mathrm{damp}}(\mathbf X^t,\dot{\mathbf X}^t;\boldsymbol\eta)+\mathbf f_i^{\mathrm{int}}(\mathbf X^t;\boldsymbol\eta)+m_i\mathbf g$. Here, $\mathbf X^t$ stacks all node positions, while $m_i$, $\mathbf g$, and the three force terms denote node mass, gravity, and elastic, damping, and interaction forces, respectively. Starting from the observed initial state, we explicitly integrate with time step $\Delta t$: $\mathbf v_i^{t+\Delta t}=\mathbf v_i^t+\Delta t\,\mathbf f_i^t/m_i$ and $\mathbf x_i^{t+\Delta t}=\mathbf x_i^t+\Delta t\,\mathbf v_i^{t+\Delta t}$, yielding the differentiable trajectory $\widehat{\mathbf X}(\theta)=\operatorname{Simulate}(\mathcal G,\mathbf Y,\boldsymbol\eta)$.

% We supervise the simulated trajectories with the recovered 3D motion.
% After physics fitting, we optimize the canonical Gaussian appearance
% parameters $\varphi$ using rendering supervision, while keeping the
% fitted physics fixed.
% The Gaussians are bound to the original physical nodes in the
% reference configuration, as detailed in Appendix~B.6.
% Although explicit spring dynamics are computationally efficient, their material response depends on the underlying graph structure. We next introduce response carriers to improve material expressiveness without replacing the reconstructed spring graph.

\subsection{Relaxable Response Carriers}
\label{sec:material_response}
As discussed in Sec.~\ref{sec:problem_formulation}, conventional central-force spring networks couple
volumetric and shear responses under affine deformation.
Hybrid models~\citep{golec2020hybridMSS} introduce
additional volumetric energy on regular lattices, but
such structures poorly match the irregular graphs
reconstructed from videos.

\begin{figure*}[t]
\centering
\includegraphics[width=\textwidth]{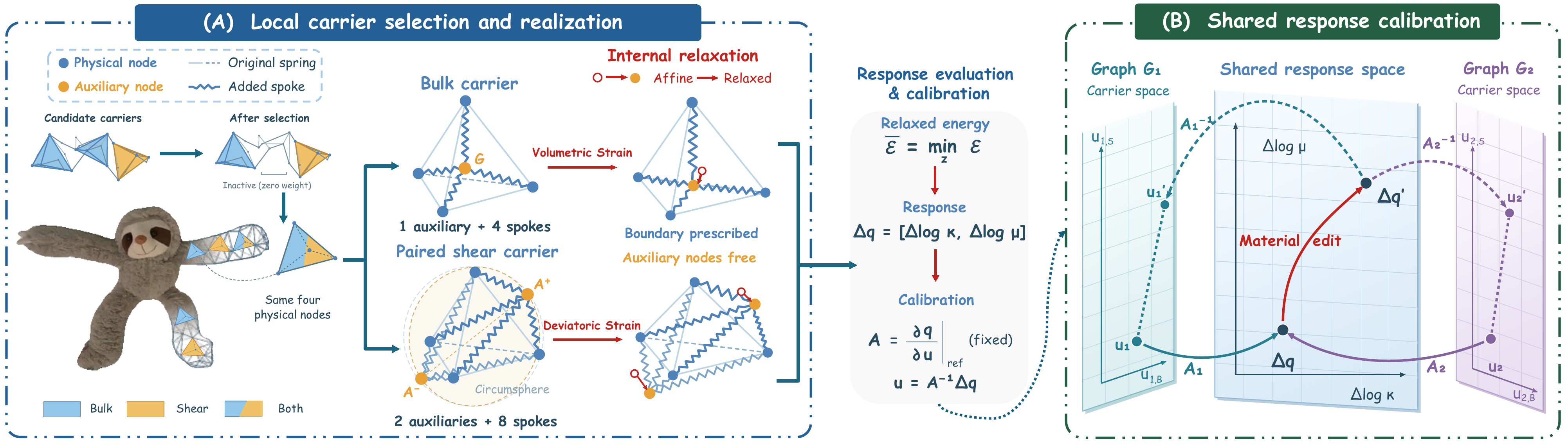}
\caption{\textbf{Response carriers and calibration.}
\underline{\textit{Part (A)}:} Candidate carriers on local four-node cells are jointly
selected to meet bulk and shear response targets while
minimizing auxiliary mass, then realized using auxiliary
nodes and spokes. Under standardized volumetric and
deviatoric strains, internal relaxation yields the
corresponding responses. 
\underline{\textit{Part (B)}:} Each graph is calibrated at its reference state to relate carrier controls to changes in material response. A shared response space then maps the same material edit to graph-specific spring updates.}
\label{fig:morphtwin-carrier}
\end{figure*}

MorphTwin instead introduces response freedom directly into the reconstructed graph $\mathcal G_0$ by augmenting selected four-node cells with relaxable internal nodes, forming $\mathcal G=(\mathcal P,\mathcal S)$ as illustrated  in Fig.~\ref{fig:morphtwin-carrier}(A).
A bulk carrier adds a centroid auxiliary node, while a paired shear carrier adds two symmetric auxiliary nodes on the circumsphere.
These candidates are geometrically screened and jointly selected to match bulk and shear response targets while minimizing auxiliary mass, as detailed in Appendix~\ref{app:carrier_construction}.
The original springs are retained, and all auxiliary nodes participate in the same explicit central-force dynamics, without requiring a regular volumetric mesh.
Each edge has a base coefficient
$Y_e(\mathbf b,\mathbf 0)$, with carrier spokes following the local backbone scale.
A fixed label
$\mathbf w_e\in\{(0,0),(1,0),(0,1)\}$
distinguishes backbone, bulk carrier, and shear carrier edges.
The controls $\mathbf u=(u_B,u_S)^\top$ adjust the corresponding spring families through logarithmic stiffness offsets:
\begin{equation}
\log Y_e(\mathbf b,\mathbf u)
=
\log Y_e(\mathbf b,\mathbf 0)
+\mathbf w_e^\top\mathbf u,
\label{eq:carrier_materialization}
\end{equation}
where $\mathbf b$ specifies backbone stiffnesses, while $u_B$ and $u_S$ independently scale
corresponding carrier spring stiffnesses.

% The key distinction from an affine central-force network lies in the internal relaxation.
% Under a prescribed small strain $\mathbf E$, the original physical nodes deform affinely, while auxiliary nodes can undergo additional nonaffine displacements $\mathbf z$.
% Minimizing over these internal degrees of freedom gives the reduced elastic energy, which is detailed in Supp. Sec~B.2:
% \begin{equation}
% \overline{\mathcal E}(\mathbf E)
% =
% \min_{\mathbf z}\frac{1}{2}
% \sum_{e=(i,j)\in\mathcal S}
% Y_eL_e
% \left[
% \mathbf n_e^\top\mathbf E\mathbf n_e
% +
% \frac{\mathbf n_e^\top(\mathbf z_j-\mathbf z_i)}{L_e}
% \right]^2.
% \label{eq:relaxed_energy}
% \end{equation}
% Here, $\mathbf z_i=\mathbf 0$ for original physical nodes, while auxiliary displacements are free.
% Unlike the fully affine setting that yields the central-force restriction $\lambda=\mu$, internal relaxation modifies the effective stiffness experienced by the physical nodes.
% The resulting energy reduction depends on both the deformation mode and carrier geometry, allowing volumetric and shear responses to vary beyond the fixed affine relation.
% The two carrier families provide distinct response sensitivities, enabling locally independent control when their response Jacobian has full rank.
% The minimization is used only for offline response evaluation. During simulation, auxiliary nodes evolve explicitly without relaxation at each step.

The affine restriction $\lambda=\mu$ in
Eq.~\eqref{eq:cauchy_limit} arises when all nodes
follow the same imposed affine deformation.
MorphTwin retains central-force springs but relaxes
this assumption for auxiliary nodes.
Under small strain $\mathbf E$, physical nodes deform
affinely ($\mathbf z_i=\mathbf 0$), while auxiliary nodes
undergo additional nonaffine displacements $\mathbf z$.
Minimizing over these displacements yields the
reduced elastic energy:
\begin{equation}
\overline{\mathcal E}(\mathbf E)
=
\min_{\mathbf z}\frac{1}{2}
\sum_{e=(i,j)\in\mathcal S}
Y_eL_e
\left[
\mathbf n_e^\top\mathbf E\mathbf n_e
+
\frac{\mathbf n_e^\top(\mathbf z_j-\mathbf z_i)}{L_e}
\right]^2.
\label{eq:relaxed_energy}
\end{equation}
The bracket combines each spring's affine strain
with a nonaffine correction due to auxiliary motion.
This relaxation can lower the energy by different
amounts for volumetric and shear deformation,
depending on carrier geometry.
Hence, the effective bulk and shear responses need
not satisfy the affine relation $\lambda=\mu$,
despite retaining central-force springs.
The two carrier families provide distinct response
sensitivities, enabling locally independent control
when the response Jacobian has full rank.
Minimization is used only for offline response
evaluation, while online simulation advances all nodes
explicitly, without per-step minimization.

\subsection{Energy Response Calibration}
\label{sec:material_identification}

To measure resistance to volume and shape changes, we evaluate
$\overline{\mathcal E}$ under unit volumetric strain
$\mathbf E_{\rm vol}=\mathbf I/\sqrt{3}$ and five orthonormal
deviatoric strains $\{\mathbf E_{\rm dev}^{(a)}\}_{a=1}^{5}$
spanning the volume-preserving strain modes in 3D:
\begin{equation}
\kappa=\frac23\overline{\mathcal E}(\mathbf E_{\rm vol}),\quad
\mu=\frac15\sum_{a=1}^{5}
\overline{\mathcal E}(\mathbf E_{\rm dev}^{(a)}),\quad
\mathbf q=[\log\kappa,\log\mu]^\top.
\label{eq:effective_response}
\end{equation}
The prefactors normalize these energies. $\kappa$ and $\mu$
are graph-integrated bulk and shear responses (dependence
on $(\mathbf b,\mathbf u)$ omitted). Their continuum moduli are
$\kappa/V$ and $\mu/V$ for reference volume $V$.
Logarithmic differences encode relative response changes,
canceling the fixed $V$ within each graph.
% Here, $\kappa$ and $\mu$ are the elastic responses
% integrated over the graph, with dependence on
% $(\mathbf b,\mathbf u)$ omitted for brevity.
% For an object with reference volume $V$, the
% corresponding continuum moduli are $\kappa/V$
% and $\mu/V$.
% Since $V$ is fixed for each graph, it cancels
% in relative logarithmic response changes.

Although $u_B$ and $u_S$ scale distinct carrier families,
each can affect both bulk and shear responses.
Their effects also depend on graph geometry and connectivity.
For each graph, let $\mathbf b^0$ denote the initial
backbone stiffness parameters used for calibration.
We define the reference response as
$\mathbf q^0=\mathbf q(\mathbf b^0,\mathbf 0)$
and compute the sensitivity matrix:
% \begin{equation}
% \begin{gathered}
% \mathbf A=
% \left.\frac{\partial\mathbf q}{\partial\mathbf u}
% \right|_{(\mathbf b^0,\mathbf 0)},\\
% \Delta\mathbf q:=
% \mathbf q(\mathbf b^0,\mathbf u)-\mathbf q^0
% \simeq\mathbf A\mathbf u,\quad
% \mathbf u=\mathbf A^{-1}\Delta\mathbf q.
% \end{gathered}
% \label{eq:response_calibration}
% \end{equation}
\begin{equation}
\mathbf A=
\frac{\partial\mathbf q}{\partial\mathbf u}
,\quad
\Delta\mathbf q:=
\mathbf q(\mathbf b^0,\mathbf u)-\mathbf q^0
\simeq\mathbf A\mathbf u,\quad
\mathbf u=\mathbf A^{-1}\Delta\mathbf q.
\label{eq:response_calibration}
\end{equation}
For full-rank $\mathbf A$, the inverse mapping
enables independent local control of bulk and
shear responses.
Each graph converts a shared relative response
change into its own spring controls.
We compute $\mathbf A$ once and keep it fixed
during fitting.

\subsection{Material Identification and Interaction}
\label{sec:editing}

\paragraph{Material identification.}
With masses fixed, we fit $\boldsymbol \theta=(\Delta\mathbf q,\mathbf b,\eta)$
and retain the damping and contact distance from initialization.
The carrier controls $\mathbf u=\mathbf A^{-1}\Delta\mathbf q$
determine the spring stiffnesses $\mathbf Y(\mathbf b,\mathbf u)$.
Explicit simulation yields
$\widehat{\mathbf X}(\boldsymbol \theta)
=\operatorname{Simulate}(\mathcal G,\mathbf Y,\eta)$,
which collects the physical-node positions over time.
We first fit physics with
$\mathcal L_{\mathrm{motion}}
=\mathcal L_{\mathrm{geo}}+\mathcal L_{\mathrm{track}}$ as geometry and 3D correspondences supervision,
then freeze $\boldsymbol \theta$ and fit Gaussian appearance $\boldsymbol \varphi$
to RGB observations:
\begin{equation}
\begin{aligned}
\min_{\boldsymbol\theta}\quad &
\mathcal L_{\rm motion}
(\widehat{\mathbf X}(\boldsymbol\theta),\mathcal V),\\
\min_{\boldsymbol\varphi}\quad &
\mathcal L_{\rm RGB}
(\operatorname{Render}(
\boldsymbol\varphi,\widehat{\mathbf X}(\boldsymbol\theta)),
\mathcal V_{\rm RGB}).
\end{aligned}
\label{eq:material_identification}
\end{equation}

% Here,
% $\mathcal L_{\mathrm{motion}}
% =\mathcal L_{\mathrm{geo}}+\mathcal L_{\mathrm{track}}$
% supervises geometry and 3D correspondences. Carrier geometry, edge labels, and $(\mathbf q^0,\mathbf A)$
% remain fixed during fitting.

% Gaussian deformation is driven by the simulated
% physical nodes through fixed local bindings,
% which is detailed in Supp. Sec.~C.6.

% \paragraph{Material editing and transfer.}
% Material edits specify a relative response change
% $\Delta\mathbf q'$, mapped locally to carrier controls
% $\mathbf u'=\mathbf A^{-1}\Delta\mathbf q'$.
% For transfer, a source object's recovered material
% response defines the target for another graph,
% which uses its own calibration rather than copying
% source spring coefficients.
% Both operations preserve the target's reference
% geometry and Gaussian appearance.
% Backbone stiffnesses $\mathbf b$ remain separately
% editable.
% New interactions are simulated explicitly without refitting.

\paragraph{Material morphing and reuse.}
Edits adjust the overall elastic scale or change response coordinates
through the local calibration.
For cross-graph reuse, the target realizes the source response through
its own offline calibration, preserving its reference geometry
and canonical appearance.

\paragraph{Interaction and fracture.}
Novel grasps drive the twin through force-limited control springs
and explicit integration.
Gaussian bindings are fixed on intact graphs, and when springs fail
irreversibly under excessive strain, appearance support uses
surviving local connections.
Nonlinear response transfer and Gaussian binding are detailed in Appendices~\ref{app:material_reuse_fracture} and~\ref{app:gaussian_binding}, respectively.

% Preamble:
% \usepackage{booktabs,adjustbox}
% \usepackage[table]{xcolor}

\begin{table*}[htbp]
\centering
\footnotesize
\color{black}

\setlength{\tabcolsep}{2.5pt}
\renewcommand{\arraystretch}{1.15}
\setlength{\heavyrulewidth}{0.45pt}
\setlength{\lightrulewidth}{0.30pt}
\setlength{\cmidrulewidth}{0.25pt}
\setlength{\aboverulesep}{0.3ex}
\setlength{\belowrulesep}{0.3ex}
\setlength{\belowcaptionskip}{4pt}

\caption{Reconstruction and future prediction on 20 PhysTwin scenes using 3 RGB-D cameras.
Simulator indicates the simulation framework used by each method.
CD and Track are in meters.
We highlight the \textbf{best} and \underline{second-best}
results for each metric and task.}
\label{tab:phystwin_reconstruction_future}

\begin{adjustbox}{max width=\linewidth}
\begin{tabular}{lc*{5}{c}@{\hspace{7pt}}*{5}{c}}
\toprule

\rowcolor{gray!10}
& &
\multicolumn{5}{c}{\textbf{Reconstruction}}
& \multicolumn{5}{c}{\textbf{Future Prediction}} \\

\cmidrule(lr){3-7}
\cmidrule(lr){8-12}

Method & Simulator
& PSNR $\uparrow$
& SSIM $\uparrow$
& LPIPS $\downarrow$
& CD $\downarrow$
& Track $\downarrow$
& PSNR $\uparrow$
& SSIM $\uparrow$
& LPIPS $\downarrow$
& CD $\downarrow$
& Track $\downarrow$ \\
\midrule

PhysTwin (ICCV'25)~\citep{jiang2025phystwin}
& Mass-Spring
& \underline{31.12}
& \underline{0.9625}
& \underline{0.0275}
& \underline{0.00553}
& 0.0094
& 25.16
& 0.9350
& 0.0610
& 0.01200
& 0.0230 \\

PGND (RSS'25)~\citep{zhang2025pgnd}
& Neural
& 27.88
& 0.9486
& 0.0518
& 0.07739
& 0.0632
& 21.57
& 0.9145
& 0.1171
& 0.23661
& 0.1988 \\

Boba (ECCV'26)~\citep{pang2026boba}
& Mass-Spring
& 31.05
& 0.9621
& 0.0322
& 0.00736
& \textbf{0.0090}
& \underline{25.92}
& \underline{0.9451}
& \underline{0.0606}
& 0.01474
& \underline{0.0223} \\

DeformMaster (NeurIPS'26)~\citep{li2026deformmaster}
& MPM + Neural
& 29.81
& 0.9534
& 0.0315
& 0.00682
& 0.0115
& 25.41
& 0.9360
& 0.0610
& \underline{0.01140}
& 0.0240 \\

MatPhys (arXiv'26)~\citep{yang2026matphys}
& Mass-Spring
& 27.39
& 0.9454
& 0.0461
& 0.01282
& 0.0211
& 23.03
& 0.9257
& 0.0855
& 0.02854
& 0.0463 \\

\rowcolor{blue!6}
\textbf{MorphTwin (Ours)}
& Mass-Spring
& \textbf{34.65}
& \textbf{0.9821}
& \textbf{0.0213}
& \textbf{0.00552}
& \underline{0.0092}
& \textbf{27.15}
& \textbf{0.9547}
& \textbf{0.0459}
& \textbf{0.01130}
& \textbf{0.0221} \\

\bottomrule
\end{tabular}
\end{adjustbox}
\end{table*}

\section{Experiments}
\label{sec:experiment}

\subsection{Experimental Setup}
\label{sec:experimental_setup}

\paragraph{Benchmarks and baselines.}
We evaluate MorphTwin on the 20 PhysTwin scenes used by
DeformMaster~\citep{li2026deformmaster}, 14 synthetic
Spring-Gaus~\citep{zhong2024springgaussian} development scenes
with known material parameters, and our self-captured egocentric interaction benchmark comprising 15 spatiotemporally aligned
sequences with stereo and external RGB-D
observations (capture details in Appendix~\ref{app:dataset}).
Real-world benchmarks assess reconstruction and future prediction,
while Spring-Gaus evaluates material property identification.
Baselines include PhysTwin~\citep{jiang2025phystwin},
DeformMaster, PGND~\citep{zhang2025pgnd},
Boba~\citep{pang2026boba}, GIC~\citep{cai2024gic} and MatPhys~\citep{yang2026matphys}.
For DeformMaster, Boba and MatPhys, we use released checkpoints, Gaussian assets and code, while training PGND and GIC under matched settings.

\paragraph{Evaluation.}
Following DeformMaster~\citep{li2026deformmaster}, we report
foreground-masked PSNR~\citep{hore2010psnr_ssim},
SSIM~\citep{wang2004ssim}, and LPIPS~\citep{zhang2018lpips}.
We also report Chamfer distance~\citep{wu2021chamfer_distance},
tracking error, relative Young's modulus error, and absolute
Poisson's ratio error.
All experiments use a single NVIDIA GeForce RTX 4090 GPU.
Further details are provided in Appendix~\ref{app:experiments}.

\subsection{Quantitative Evaluation}
\label{sec:phystwin_quality_efficiency}

\paragraph{Reconstruction and future prediction.}
Table~\ref{tab:phystwin_reconstruction_future} reports reconstruction
and future prediction results on the PhysTwin benchmark.
With PhysTwin's preprocessing unchanged, MorphTwin supports
calibrated material control and achieves the highest PSNR and SSIM
and the lowest LPIPS among the evaluated methods in both settings,
with PSNR gains of 3.53~dB for reconstruction and 1.99~dB for
future prediction over PhysTwin.
LPIPS decreases by 22.5\% for reconstruction
and 24.8\% for future prediction.
Future prediction CD decreases from 12.00 to 11.30~mm,
and tracking error decreases from 23.0 to 22.1~mm.

% Preamble:
% \usepackage{booktabs,multirow,graphicx}
% \usepackage[table]{xcolor}
% \usepackage{arydshln} % Load after xcolor

\begin{table}[t]
\centering
\footnotesize
\color{black}

\setlength{\tabcolsep}{1pt}
\renewcommand{\arraystretch}{1.05}
\setlength{\heavyrulewidth}{0.45pt}
\setlength{\lightrulewidth}{0.30pt}
\setlength{\aboverulesep}{0.20ex}
\setlength{\belowrulesep}{0.20ex}
\setlength{\belowcaptionskip}{3pt}

\caption{Quantitative evaluation on 15-sequence benchmark using
single RGB-D, egocentric stereo, and combined views.
CD and Track are in meters.
\textbf{Best}/\underline{second-best} per input setting and task.}
\label{tab:view_configuration_ablation}

\begin{tabularx}{\linewidth}
{@{}cl>{\raggedright\arraybackslash}Xccccc@{}}
\toprule
& Views & Method
& PSNR$\uparrow$
& SSIM$\uparrow$
& LPIPS$\downarrow$
& CD$\downarrow$
& Track$\downarrow$ \\
\midrule

% ================= Reconstruction =================

% ---------- All Views ----------
& & PhysTwin~\citep{jiang2025phystwin}
& \underline{28.74} & \underline{0.9451}
& \underline{0.0598} & \underline{0.00946}
& \underline{0.0172} \\

& & PGND~\citep{zhang2025pgnd}
& 27.59 & 0.9403 & 0.0722 & 0.01448 & 0.0192 \\

& & GIC~\citep{cai2024gic}
& 24.58 & 0.9225 & 0.1302 & 0.03457 & 0.0756 \\

\rowcolor{blue!6}
\cellcolor{white}
& \cellcolor{white}\multirow{-4}{*}{All}
& \textbf{MorphTwin}
& \textbf{32.36} & \textbf{0.9665}
& \textbf{0.0511} & \textbf{0.00925}
& \textbf{0.0171} \\

\cdashline{2-8}[1pt/1.5pt]

% ---------- Egocentric ----------
& & PhysTwin~\citep{jiang2025phystwin}
& \underline{28.06} & \underline{0.9395}
& \underline{0.0782} & \underline{0.00952}
& \underline{0.0178} \\

& & PGND~\citep{zhang2025pgnd}
& 27.18 & 0.9342 & 0.0893 & 0.01416 & 0.0196 \\

& & GIC~\citep{cai2024gic}
& 24.44 & 0.9237 & 0.1288 & 0.03252 & 0.0737 \\

\rowcolor{blue!6}
\cellcolor{white}
& \cellcolor{white}\multirow{-4}{*}{Egocentric}
& \textbf{MorphTwin}
& \textbf{31.13} & \textbf{0.9531}
& \textbf{0.0669} & \textbf{0.00941}
& \textbf{0.0176} \\

\cdashline{2-8}[1pt/1.5pt]

% ---------- Single RGB-D ----------
& & PhysTwin~\citep{jiang2025phystwin}
& 26.38 & 0.9291 & 0.0887
& \textbf{0.01002} & 0.0220 \\

& & PGND~\citep{zhang2025pgnd}
& \underline{26.47} & \underline{0.9377}
& \underline{0.0779} & 0.01303
& \underline{0.0191} \\

& & GIC~\citep{cai2024gic}
& 22.54 & 0.9183 & 0.1334 & 0.03228 & 0.0734 \\

\rowcolor{blue!6}
\cellcolor{white}
\multirow{-12}{*}{\rotatebox[origin=c]{90}{Reconstruction}}
& \cellcolor{white}
  \multirow{-4}{*}{\shortstack[l]{Single\\RGB-D}}
& \textbf{MorphTwin}
& \textbf{27.37} & \textbf{0.9451}
& \textbf{0.0772} & \underline{0.01078}
& \textbf{0.0189} \\

\midrule

% ================= Future Prediction =================

% ---------- All Views ----------
& & PhysTwin~\citep{jiang2025phystwin}
& \underline{25.27} & \underline{0.9321}
& \underline{0.0850} & \underline{0.01597}
& \underline{0.0322} \\

& & PGND~\citep{zhang2025pgnd}
& 22.19 & 0.9103 & 0.1418 & 0.06808 & 0.0798 \\

& & GIC~\citep{cai2024gic}
& 20.85 & 0.9023 & 0.1652 & 0.05989 & 0.1509 \\

\rowcolor{blue!6}
\cellcolor{white}
& \cellcolor{white}\multirow{-4}{*}{All}
& \textbf{MorphTwin}
& \textbf{26.78} & \textbf{0.9460}
& \textbf{0.0760} & \textbf{0.01492}
& \textbf{0.0306} \\

\cdashline{2-8}[1pt/1.5pt]

% ---------- Egocentric ----------
& & PhysTwin~\citep{jiang2025phystwin}
& \underline{25.04} & \underline{0.9291}
& \underline{0.0971} & \underline{0.01635}
& \underline{0.0331} \\

& & PGND~\citep{zhang2025pgnd}
& 22.26 & 0.9093 & 0.1434 & 0.06265 & 0.0757 \\

& & GIC~\citep{cai2024gic}
& 21.02 & 0.9013 & 0.1634 & 0.05586 & 0.1392 \\

\rowcolor{blue!6}
\cellcolor{white}
& \cellcolor{white}\multirow{-4}{*}{Egocentric}
& \textbf{MorphTwin}
& \textbf{26.36} & \textbf{0.9364}
& \textbf{0.0882} & \textbf{0.01508}
& \textbf{0.0317} \\

\cdashline{2-8}[1pt/1.5pt]

% ---------- Single RGB-D ----------
& & PhysTwin~\citep{jiang2025phystwin}
& \underline{23.80} & \underline{0.9202}
& \underline{0.1087} & \underline{0.01694}
& \underline{0.0376} \\

& & PGND~\citep{zhang2025pgnd}
& 21.90 & 0.9114 & 0.1374 & 0.06071 & 0.0747 \\

& & GIC~\citep{cai2024gic}
& 19.82 & 0.8967 & 0.1661 & 0.05660 & 0.1392 \\

\rowcolor{blue!6}
\cellcolor{white}
\multirow{-12}{*}{\rotatebox[origin=c]{90}{Future Prediction}}
& \cellcolor{white}
  \multirow{-4}{*}{\shortstack[l]{Single\\RGB-D}}
& \textbf{MorphTwin}
& \textbf{24.41} & \textbf{0.9288}
& \textbf{0.0955} & \textbf{0.01680}
& \textbf{0.0322} \\

\bottomrule
\end{tabularx}
\end{table}

Table~\ref{tab:view_configuration_ablation} compares methods on our
15-sequence egocentric benchmark using three view modes: single RGB-D,
egocentric stereo, and all views.
With all views, MorphTwin reduces reconstruction and future-prediction
LPIPS by 14.6\% and 10.6\% over PhysTwin, respectively.
Moving from single RGB-D to all views increases reconstruction PSNR
by 4.99~dB for MorphTwin versus 2.36~dB for PhysTwin.
MorphTwin also outperforms PhysTwin on all metrics with egocentric
stereo alone, demonstrating its advantage in real-world reconstruction
and future interaction prediction.

\begin{figure}[t]
\centering
\includegraphics[width=0.47\textwidth]{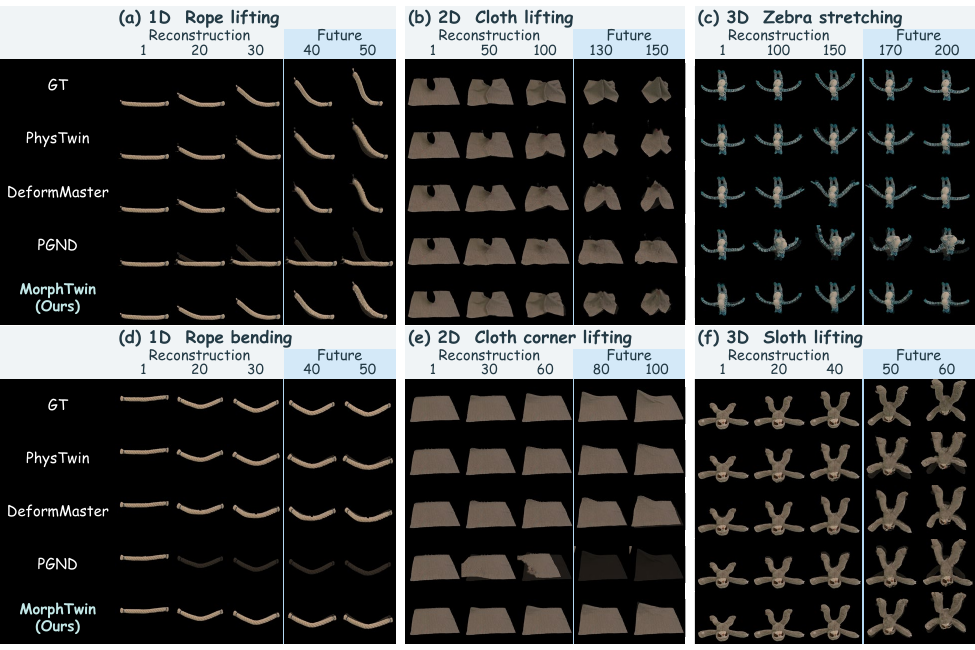}
\caption{Reconstruction and prediction on PhysTwin scenes.
PGND can exhibit unstable future predictions, while PhysTwin's
explicit mass-spring model may not fully reproduce material responses.
Semi-transparent GT overlays highlight discrepancies.}
\label{fig:phystwin-reconstruction-future}
\end{figure}

\paragraph{Efficient explicit simulation.}
MorphTwin calibrates volumetric and shape response targets into spring parameters for each graph offline. Online simulation directly
advances the calibrated explicit graph, without per-step material
identification, energy alignment, or an additional material-prediction
network.
Table~\ref{tab:dimension_efficiency} shows that MorphTwin achieves
higher offline and same-node interactive throughput than PhysTwin
and DeformMaster across all three dimensional groups.  In addition, MorphTwin realizes material responses through an explicit spring graph,
making it compatible with complementary acceleration strategies developed
for PhysTwin.

\begin{figure*}[t]
\centering
\includegraphics[width=0.9\linewidth]
{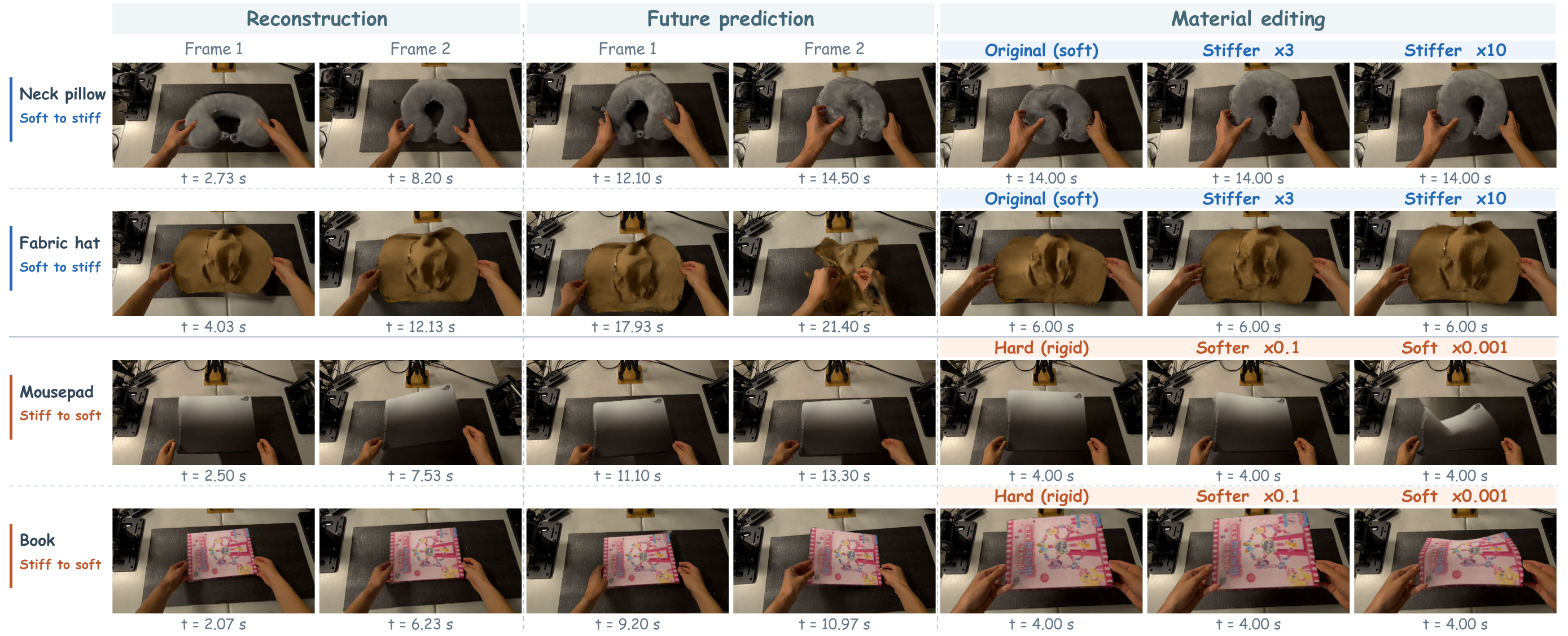}
\caption{
\textbf{Material editing on our egocentric interaction benchmark.}
Four representative sequences from our 15-sequence benchmark
show reconstruction, future prediction, and material editing
under the same hand interactions.
At the same observation time within each row,
stiffening the pillow and hat reduces deformation,
while softening the mousepad and book enables bending.
}
\label{fig:ego_material_editing}
\end{figure*}

% Requires booktabs, multirow, adjustbox, and xcolor with [table].
\begin{table}[t]
\centering
\footnotesize
\setlength{\tabcolsep}{3.5pt}
\renewcommand{\arraystretch}{1.05}
\setlength{\aboverulesep}{0.35ex}
\setlength{\belowrulesep}{0.35ex}
\caption{Physics optimization cost and inference efficiency on the 20 PhysTwin
scenes. Bold indicates the highest reported FPS and lowest reported memory
per group.}
\label{tab:dimension_efficiency}

\begin{adjustbox}{max width=\linewidth}
\begin{tabular}{@{}llrrrr@{}}
\toprule
\textbf{Scene type} & \textbf{Method}
& \multicolumn{1}{c}{\shortstack{Training $\downarrow$\\(min/scene)}}
& \multicolumn{1}{c}{\shortstack{Offline\\FPS $\uparrow$}}
& \multicolumn{1}{c}{\shortstack{Interactive\\FPS $\uparrow$}}
& \multicolumn{1}{c}{\shortstack{Memory $\downarrow$\\(GiB)}} \\
\midrule
\multirow{3}{*}{\shortstack[l]{\textbf{1D curves}\\[-1pt]\scriptsize(3 scenes)}}
& PhysTwin~\citep{jiang2025phystwin} & 4.78 & 85.00 & 50.68 & 1.01 \\
& DeformMaster~\citep{li2026deformmaster} & N/A & 37.42 & 28.36 & \textbf{0.81} \\
& \cellcolor{blue!6}\textbf{MorphTwin}
& \cellcolor{blue!6}10.82
& \cellcolor{blue!6}\textbf{106.99}
& \cellcolor{blue!6}\textbf{62.87}
& \cellcolor{blue!6}1.12 \\
\midrule
\multirow{3}{*}{\shortstack[l]{\textbf{2D surfaces}\\[-1pt]\scriptsize(9 scenes)}}
& PhysTwin~\citep{jiang2025phystwin} & 72.05 & 24.72 & 19.47 & \textbf{1.79} \\
& DeformMaster~\citep{li2026deformmaster} & N/A & 22.35 & 18.25 & 2.18 \\
& \cellcolor{blue!6}\textbf{MorphTwin}
& \cellcolor{blue!6}109.20
& \cellcolor{blue!6}\textbf{26.26}
& \cellcolor{blue!6}\textbf{21.47}
& \cellcolor{blue!6}2.02 \\
\midrule
\multirow{3}{*}{\shortstack[l]{\textbf{3D solids}\\[-1pt]\scriptsize(8 scenes)}}
& PhysTwin~\citep{jiang2025phystwin} & 7.60 & 47.75 & 33.77 & 1.65 \\
& DeformMaster~\citep{li2026deformmaster} & N/A & 18.55 & 15.63 & 3.59 \\
& \cellcolor{blue!6}\textbf{MorphTwin}
& \cellcolor{blue!6}16.32
& \cellcolor{blue!6}\textbf{67.86}
& \cellcolor{blue!6}\textbf{45.98}
& \cellcolor{blue!6}\textbf{1.44} \\
\bottomrule
\end{tabular}
\end{adjustbox}
\end{table}

\subsection{Qualitative Evaluation}
\label{sec:qualitative_evaluation}

Figure~\ref{fig:phystwin-reconstruction-future} compares MorphTwin
with PhysTwin, DeformMaster, and PGND on 1D curves, 2D surfaces,
and 3D solids using semi-transparent GT overlays.
Especially in the illustrated rope-lifting and sloth-lifting cases,
MorphTwin more closely matches the reference contours
and deformation in future frames than PhysTwin,
suggesting better generalization beyond observed frames.
Figure~\ref{fig:morphtwin-per-scene} in Appendix~\ref{app:experiments}
extends the comparison to all 15 sequences
of our benchmark.

\subsection{Downstream Applications}
\label{sec:downstream_applications}

\paragraph{Material property identification.}

\label{sec:material_recovery}

\begin{table}[t]
\centering
\scriptsize
\color{black}

\setlength{\tabcolsep}{1.5pt}
\renewcommand{\arraystretch}{1.05}
\setlength{\heavyrulewidth}{0.45pt}
\setlength{\lightrulewidth}{0.30pt}
\setlength{\aboverulesep}{0.15ex}
\setlength{\belowrulesep}{0.15ex}

\caption{Material property identification and appearance
evaluation on Spring-Gaus~\citep{zhong2024springgaussian}. $E$ error is relative (\%) and
$\nu$ error is absolute.}
\label{tab:morphtwin_main}

% ---------- Per-scene Poisson ratios ----------
\begin{adjustbox}{max width=\linewidth}
\begin{tabular}{
@{}lcc
>{\columncolor{blue!6}\bfseries}c
@{\hspace{5pt}}
lcc
>{\columncolor{blue!6}[\tabcolsep][0pt]\bfseries}c@{}
}
\toprule
\rowcolor{gray!12}[0pt][0pt]
\multicolumn{8}{@{}l@{}}{
\textbf{Poisson's Ratio Identification}} \\
Scene & $\nu_{\mathrm{GT}}$ & PhysTwin~\citep{jiang2025phystwin} & Ours
& Scene & $\nu_{\mathrm{GT}}$ & PhysTwin~\citep{jiang2025phystwin} & Ours \\
\midrule
apple    & 0.01 & 0.2492 & 0.0260
& droplet    & 0.30 & 0.2511 & 0.3268 \\
banana   & 0.30 & 0.2466 & 0.2942
& duck       & 0.30 & 0.2471 & 0.2693 \\
C        & 0.10 & 0.2445 & 0.1171
& E          & 0.40 & 0.2378 & 0.3408 \\
chess    & 0.40 & 0.2478 & 0.3662
& rope       & 0.30 & 0.2414 & 0.2939 \\
cream    & 0.30 & 0.2489 & 0.2950
& toothpaste & 0.40 & 0.2448 & 0.3757 \\
cross    & 0.30 & 0.2468 & 0.2623
& torus      & 0.30 & 0.2405 & 0.2887 \\
doraemon & 0.30 & 0.2473 & 0.3475
& V          & 0.30 & 0.2441 & 0.2961 \\
\bottomrule
\end{tabular}
\end{adjustbox}

\par\vspace{3pt}

% ---------- Transposed overall comparison ----------
\begin{adjustbox}{max width=\linewidth}
\begin{tabular}{@{}l*{8}{c}@{}}
\toprule
& \multicolumn{2}{c}{Parameter Errors}
& \multicolumn{3}{c}{Reconstruction}
& \multicolumn{3}{c@{}}{Future Prediction} \\
\cmidrule(lr){2-3}
\cmidrule(lr){4-6}
\cmidrule(l){7-9}
Method
& $E\downarrow$ & $\nu\downarrow$
& PSNR$\uparrow$ & SSIM$\uparrow$ & LPIPS$\downarrow$
& PSNR$\uparrow$ & SSIM$\uparrow$ & LPIPS$\downarrow$ \\
\midrule
Spring-Gaus
& 335.29 & 0.1189
& 21.10 & 0.9168 & 0.1130
& 15.80 & 0.8586 & 0.1733 \\
PhysTwin
& 18.22 & 0.0957
& 22.00 & 0.9286 & 0.1041
& 16.28 & 0.8720 & 0.1595 \\
\rowcolor{blue!6}[0pt][0pt]
\textbf{MorphTwin}
& \textbf{14.20} & \textbf{0.0232}
& \textbf{22.18} & \textbf{0.9296} & \textbf{0.1024}
& \textbf{16.35} & \textbf{0.8746} & \textbf{0.1554} \\
\bottomrule
\end{tabular}
\end{adjustbox}
\end{table}

We evaluate material property identification on 14 synthetic
Spring-Gaus development scenes with known Young's modulus $E$
and Poisson's ratio $\nu$.
PhysTwin and MorphTwin both optimize graph-specific spring parameters
during inverse fitting, while MorphTwin uses calibrated bulk and shear
response coordinates.
For RGB-supervised Spring-Gaus, equivalent $E$ and $\nu$ are computed
from its fitted spring network.

Table~\ref{tab:morphtwin_main} shows that MorphTwin reduces mean
absolute $\nu$ error by 75.8\% (0.0957 to 0.0232) and mean relative
$E$ error by 22.1\% (18.22\% to 14.20\%) over PhysTwin,
with lower $\nu$ error on all 14 scenes.
PhysTwin's estimates cluster within $0.2378$--$0.2511$ despite
ground truths spanning $0.01$--$0.40$, consistent with the Cauchy
restriction $\lambda=\mu$ and $\nu=1/4$ for unstressed, isotropic
3D central-force networks under affine small strains.
MorphTwin estimates $\nu=0.0260$ for \textit{apple} and $0.3757$
for \textit{toothpaste}, against ground truths of $0.01$ and $0.40$,
demonstrating identification on both sides of $1/4$.
These results support material identification for subsequent
editing and transfer, alongside improved reconstruction
and future prediction.

\paragraph{Material editing and reusable physical assets.}
\label{sec:editable_reusable_twins}

Figure~\ref{fig:ego_material_editing} demonstrates material editing
in egocentric interactions.
Under fixed interaction inputs, stiffening the neck pillow and
fabric hat reduces deformation, while softening the mousepad
and book allows bending.
These edits retain reference geometry and appearance, allowing
identified material responses to control mechanical behavior
without refitting.

Appendix~\ref{app:experiments} evaluates material transfer across objects and
graph discretizations through graph-specific calibration,
preserving targeted bulk and shear responses.
It also demonstrates reuse across interactions and progressive
tearing with connectivity-aware appearance updates.
Figure~\ref{fig:isaac_sim_interaction} shows keyboard control of reconstructed assets
and robotic interaction after import into a physics engine.

\subsection{Ablation Studies}
\label{sec:ablation}

Ablations examine questions motivated by our
core physical observations:
\textit{(a) Why decouple bulk and shear responses?}
\textit{(b) Why focus identification on elastic responses?}
\textit{(c) What enables reliable material identification and transfer?}
We also examine how stiffness constraints affect motion fitting.

\begin{figure}[t]
\centering
\includegraphics[width=0.95\linewidth]
{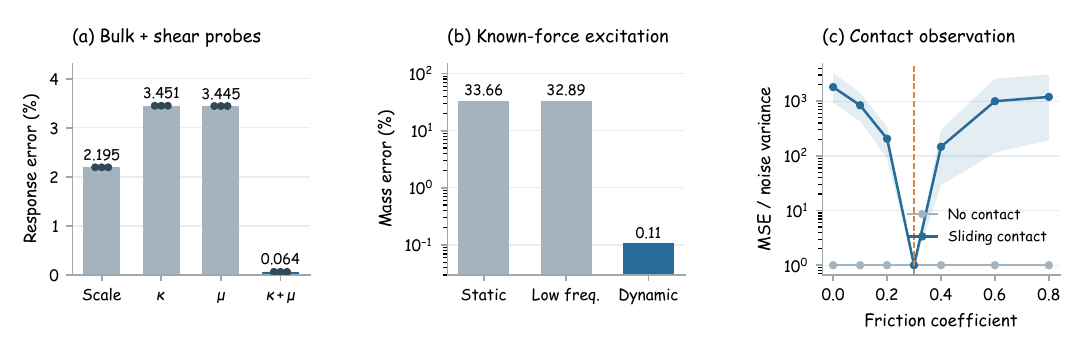}
\caption{
\textbf{Elastic parameterization and observability.}
(a) Volumetric and shear response fitting with one or two elastic controls.
(b) Mass identification under known-force excitation.
(c) Friction loss without contact and during sliding contact.
}
\label{fig:observability_ablation}
\end{figure}

\paragraph{Material parameterization and observability.}
We examine why elasticity needs two controls, why mass is not
jointly identified by default, and why friction is treated separately.
Figure~\ref{fig:observability_ablation}(a) shows that joint
$\kappa$--$\mu$ control achieves 0.064\% response error versus
2.195\% for the best single-parameter alternative,
supporting independent control of volume and shape.
Under known forces, Figure~\ref{fig:observability_ablation}(b)
shows mass error decreasing from 33.66\% under static loading
to 0.11\% under dynamic excitation, demonstrating the importance
of inertial information.
Figure~\ref{fig:observability_ablation}(c) shows a flat friction
loss without contact and a distinct minimum during sliding,
highlighting the need for informative contact.

When reconstructing interactive physical twins from sparse real-world
views, these observational requirements are often unmet,
making true mass and friction values difficult to identify reliably.
We therefore fix density and focus on identifying bulk and shear responses.

\begin{figure}[t]
\centering
\includegraphics[width=0.9\linewidth]
{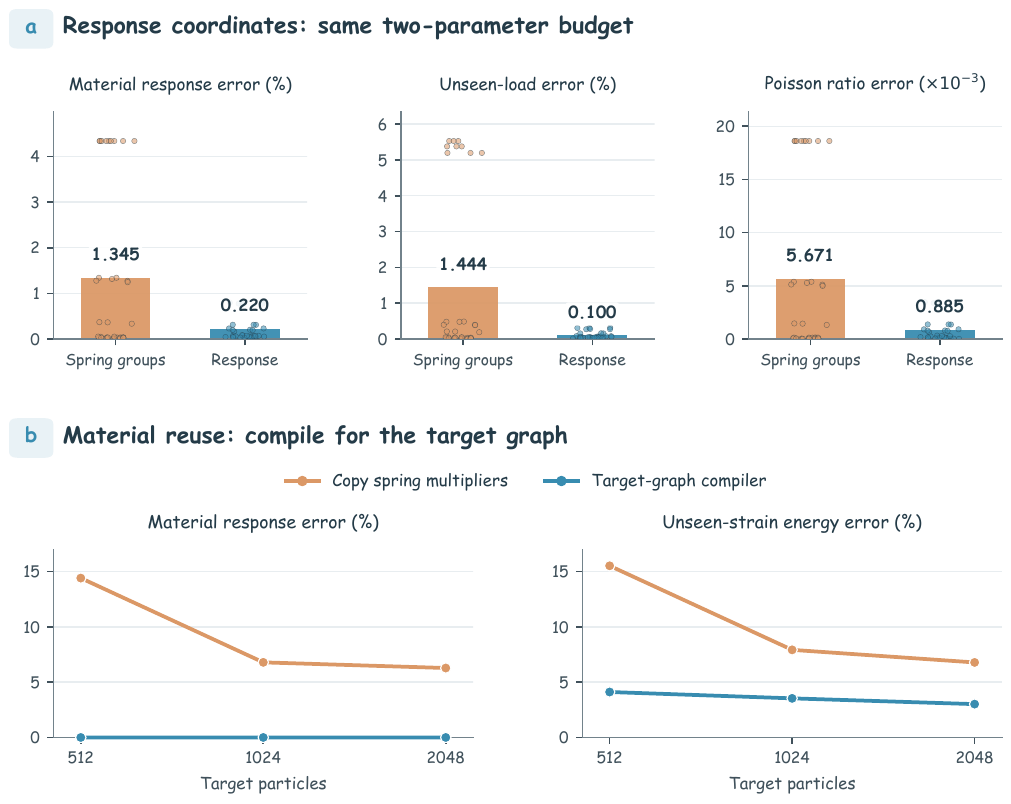}
\caption{
\textbf{Response coordinates and graph calibration.}
(a) Response coordinates improve fitting accuracy and generalization
to unseen loads under matched two-parameter budgets.
(b) Target-graph calibration reduces response and unseen-strain
energy errors compared with copied spring multipliers.
}
\label{fig:alignment_transfer_ablation}
\end{figure}

\paragraph{Response coordinates and graph calibration.}
With matched initialization, attainable ranges, and optimization budgets,
Figure~\ref{fig:alignment_transfer_ablation}(a) shows that response
coordinates reduce fitting error from 1.345\% to 0.220\%
and unseen-load error from 1.444\% to 0.100\%.
This supports an optimization advantage at fixed expressive capacity.
In the mass-normalized transfer test of
Figure~\ref{fig:alignment_transfer_ablation}(b),
target-graph calibration reduces response error from 9.5173\%
to 0.000503\% and unseen-strain energy error from 10.4833\%
to 3.6203\%, without fitting target motion.
Shared response targets enable material transfer across graphs
through graph-specific spring realizations.
The reduced energy error under unseen strains demonstrates that
calibration improves mechanical consistency beyond the calibration probes.

\begin{table}[t]
\centering
\footnotesize
\setlength{\tabcolsep}{4pt}
\caption{\textbf{Stiffness-bound scheduling.}
Comparison of fixed and scheduled
stiffness lower bounds on our egocentric interaction benchmark
under all-view supervision.
CD and Track are in meters.}
\label{tab:stiffness_bound_ablation_local}
\begin{tabular}{lcccc}
\toprule
& \multicolumn{2}{c}{Reconstruction}
& \multicolumn{2}{c}{Future prediction} \\
\cmidrule(lr){2-3}\cmidrule(lr){4-5}
Lower bound & CD$\downarrow$ & Track$\downarrow$
& CD$\downarrow$ & Track$\downarrow$ \\
\midrule
Fixed
& 0.01050 & 0.0188
& 0.01509 & \textbf{0.0301} \\
\rowcolor{blue!6}
\textbf{Scheduled}
& \textbf{0.00925} & \textbf{0.0171}
& \textbf{0.01492} & 0.0306 \\
\bottomrule
\end{tabular}
\end{table}

\paragraph{Physical parameter refinement.}
During early optimization, overly soft springs may
fit motion discrepancies at the expense of meaningful
material responses. We therefore bound each backbone spring stiffness
by $k_e \ge k_{e,\min}=0.5k_e^{(0)}$, where CMA provides
the initial stiffness $k_e^{(0)}$. The fixed variant retains this bound throughout fitting.
The scheduled variant sets $k_{e,\min}L_e=1$ in simulator force units after 100 epochs, in order to allow finer deformation fitting.
Table~\ref{tab:stiffness_bound_ablation_local} shows that scheduling
reduces reconstruction CD and tracking errors by
11.9\% and 9.0\%, respectively.

\section{Conclusion and Future Work}
\label{sec:conclusion}

We presented MorphTwin for accurate material recovery, interpretable
editing, and reuse across objects from videos while retaining efficient
explicit simulation. Experiments demonstrate improved reconstruction
and future prediction, including from egocentric observations.
These results support building physical twins with reusable,
explicitly controllable material behavior.
Our current formulation focuses on isotropic elasticity with fixed
density. Future work will integrate feed-forward reconstruction
to streamline physical twin creation and broaden its applications.

% \section*{LLM Usage Statement}
% \label{app:ai_use}

% For manuscript preparation, we used ChatGPT (GPT-6) to improve the grammar, wording, readability, and internal consistency of author-written text. All AI-assisted edits were manually reviewed and revised by the authors. 

% We also used Codex (GPT-6 Astra) as an AI agent to assist with processing our self-captured egocentric stereo recordings from Meta-Quest-3 VR devices and RGB-D videos from depth cameras, primarily for keypoint annotation and temporal alignment. Every processing stage underwent rigorous and systematic human review. The authors checked the resulting annotations and temporal correspondences and corrected them where necessary before use.

{
    \small
    \bibliographystyle{ieeenat_fullname}
    \bibliography{main}
}

%%%%%%%%% APPENDIX
\clearpage
\appendix
\twocolumn[{%
    \centering
    \pdfbookmark[0]{Appendix}{appendix}%
    {\Large\bfseries Appendix\par}%
    \vspace{1.0em}%
}]
\appendix
\paragraph{Overview.}
This appendix provides theoretical analysis and derivations
(Sec.~\ref{app:theory}), optimization and inference details
(Sec.~\ref{app:implementation}), dataset and capture details
(Sec.~\ref{app:dataset}), experimental protocols and additional
results (Sec.~\ref{app:experiments}), and an overview of the
interactive interface and operating workflow (Sec.~\ref{app:ui}).
We provide two supplementary videos demonstrating MorphTwin's
reconstruction, interactive simulation, and material control
(Table~\ref{tab:supplementary_videos}).

\begin{table}[htbp]
\centering
\small
\setlength{\tabcolsep}{5pt}
\renewcommand{\arraystretch}{1.12}
\caption{Supplementary videos and their purposes.}
\label{tab:supplementary_videos}
\begin{tabularx}{\linewidth}{
@{}>{\raggedright\arraybackslash}p{0.38\linewidth}
>{\raggedright\arraybackslash}X@{}
}
\toprule
\textbf{Video} & \textbf{Purpose} \\
\midrule
\textbf{Video 1:}\par MorphTwin benchmark and applications
& Compare MorphTwin with PhysTwin across all scenes in MorphTwin egocentric benchmark, and demonstrate physics engine integration and
keyboard interaction. \\

\textbf{Video 2:}\par PhysTwin benchmark and material editing
& Compare MorphTwin, PhysTwin, and PGND on selected PhysTwin
benchmark scenes, and visualize material editing results. \\
\bottomrule
\end{tabularx}
\end{table}

% \section{AI Use Statement}
% \label{app:ai_use}

% For manuscript preparation, we used AI writing assistants to improve
% the grammar, wording, readability, and internal consistency of
% author-written text.
% All AI-assisted edits were manually reviewed and revised by the authors.
% We also used AI agents to assist with processing our self-captured
% egocentric stereo recordings from VR devices and RGB-D videos from
% depth cameras, primarily for keypoint annotation and temporal alignment.
% Every processing stage underwent rigorous and systematic human review.
% The authors checked the resulting annotations and temporal
% correspondences and corrected them where necessary before use.

\section{Comparison with Related Work}
\label{app:clarify}

\paragraph{Task and challenges.}
MorphTwin aims not only to reconstruct interactive physical twins
from sparse videos, but also to endow them with physically
meaningful material properties.
This requires identifying material responses from limited
deformation observations to enable generalization to novel forces
and manipulations, material editing, and reuse.
Achieving these goals at interactive speeds is challenging because
spring models can couple bulk and shear responses, while their
effective material behavior depends on graph geometry and
connectivity.

We compare representative methods through their inputs,
intermediate processing, and outputs, grouping them by their
modeling approaches and material representations.

\paragraph{Static assets and images.}
PhysGaussian~\citep{xie2024physgaussian} uses MPM~\citep{jiang2016mpm} to animate static
Gaussian~\citep{kerbl20233dgs} assets with prescribed material parameters.
PhysDreamer~\citep{zhang2024physdreamer} and
OmniPhysGS~\citep{lin2025omniphysgs} use generative motion priors to
infer material behavior from static inputs, while
PhysGen3D~\citep{chen2025physgen3d} predicts geometry and physical
properties from one image.
WonderPlay~\citep{li2025wonderplay} and
PerpetualWonder~\citep{zhan2026perpetualwonder} combine simulation
with generative refinement for action-conditioned dynamics.
Static appearance underconstrains object-specific material
responses.
MorphTwin uses observed interaction and deformation to constrain
material identification.

\paragraph{Video-based physical reconstruction.}
PAC-NeRF~\citep{li2023pac}, GIC~\citep{cai2024gic}, and
MonoPhysics~\citep{rho2026monophysics} recover geometry and continuum
parameters from multi-view or monocular videos through
differentiable MPM and visual supervision, yielding models for
resimulation.
Spring-Gaus~\citep{zhong2024springgaussian},
PhysTwin~\citep{jiang2025phystwin}, and
PhysHanDI~\citep{lee2026physhandi} fit spring-based models from
multi-view videos or sparse-view RGB-D interactions for prediction
and interaction.
Boba~\citep{pang2026boba} starts from existing Gaussian twins and
accelerates rollouts using compact spring surrogates and optimized
execution.
Continuum simulation can be costly at fine resolutions, while
central-force springs couple bulk and shear in unstressed isotropic
3D networks under affine small strains.
MorphTwin enriches local deformation modes with internally
relaxable response carriers for independent local bulk and shear
control.
These carriers operate on irregular reconstructed graphs, whereas
the hybrid spring-mass model~\citep{golec2020hybridMSS} uses regular
3D cubic lattices.
Energy minimization for response evaluation and graph calibration
are performed offline, preserving efficient online dynamics with
standard central-force springs and explicit integration.

\paragraph{Learned dynamics and material parameters.}
GaussianPrediction~\citep{zhao2024gaussianprediction} predicts
keypoint motion from videos, while PGND~\citep{zhang2025pgnd}
learns velocity fields from sparse-view RGB-D interactions.
DeformMaster~\citep{li2026deformmaster} combines MPM with neural
residuals for interactive prediction and material variation.
MatPhys~\citep{yang2026matphys} and EgoPhys~\citep{kim2026egophys}
use codebooks to predict spring parameters from monocular and
egocentric RGB videos, respectively, while
NeuSpring~\citep{xu2026neuspring} fits neural spring fields to
interaction videos.
Reusing learned spring parameters requires accounting for graph
geometry and connectivity~\citep{lloyd2008identification}.
MorphTwin enables explicit material identification and control
by recovering bulk and shear responses from observed deformation
and mapping desired relative response changes to graph-specific
spring parameters through energy-based calibration.

\section{Theoretical Analysis and Derivations}
\label{app:theory}

\subsection{Bulk and Shear Coupling}
\label{app:central_coupling}

We provide the derivation of the bulk and shear coupling stated in Eq.~\eqref{eq:central_vs_iso}. Consider an unstressed three-dimensional central-force network under a symmetric small strain $\boldsymbol{\epsilon}$. For an edge with reference direction $\mathbf n$, the first-order axial strain is $\mathbf n^\top \boldsymbol{\epsilon}\mathbf n$. Under an isotropic effective response, the stiffness-weighted edge directions satisfy the standard fourth-order directional average
\begin{equation}
\left\langle
n_\alpha n_\beta n_\gamma n_\delta
\right\rangle
=
\frac{1}{15}
\left(
\delta_{\alpha\beta}\delta_{\gamma\delta}
+
\delta_{\alpha\gamma}\delta_{\beta\delta}
+
\delta_{\alpha\delta}\delta_{\beta\gamma}
\right).
\label{eq:directional_average}
\end{equation}
Here, $\alpha,\beta,\gamma,\delta\in\{1,2,3\}$
are Cartesian indices, and
$\delta_{\alpha\beta}$ is the Kronecker delta,
equal to one when $\alpha=\beta$ and zero otherwise.
The brackets denote a normalized average over
edge directions weighted by their stiffness contributions.
We also use
$\mathbf A:\mathbf B=\operatorname{tr}(\mathbf A^\top\mathbf B)
=\sum_{\alpha,\beta}A_{\alpha\beta}B_{\alpha\beta}$
to denote the double contraction. Contracting the directional average with $\epsilon_{\alpha\beta}\epsilon_{\gamma\delta}$ therefore gives
\begin{equation}
\left\langle
(\mathbf n^\top \boldsymbol{\epsilon}\mathbf n)^2
\right\rangle
=
\frac{1}{15}
\left[
(\operatorname{tr}\boldsymbol{\epsilon})^2
+
2\boldsymbol{\epsilon}:\boldsymbol{\epsilon}
\right].
\label{eq:central_directional_energy}
\end{equation}
Let $V$ denote the reference volume.
For the normalized stiffness weighting above,
the effective network stiffness density is
$c=V^{-1}\sum_{e\in\mathcal S}Y_eL_e$.
The resulting energy density is
\begin{equation}
W_{\mathrm{central}}(\boldsymbol{\epsilon})
=
\frac{c}{30}
\left[
(\operatorname{tr}\boldsymbol{\epsilon})^2
+
2\boldsymbol{\epsilon}:\boldsymbol{\epsilon}
\right].
\label{eq:central_effective_energy}
\end{equation}

To characterize which isotropic material responses this spring network can represent, we compare its effective energy with the general small-strain energy of an isotropic linear-elastic solid. The latter admits two independent elastic parameters, whereas Eq.~\eqref{eq:central_effective_energy} contains only the single network stiffness scale $c$. The isotropic constitutive law is $\boldsymbol{\sigma}=\lambda\,\operatorname{tr}(\boldsymbol{\epsilon})\mathbf I+2\mu\boldsymbol{\epsilon}$, where $\boldsymbol{\sigma}$ is the Cauchy stress tensor
and $\mathbf I\in\mathbb R^{3\times3}$ is the
three-dimensional identity matrix.
The corresponding strain-energy density~\citep{reddy2013introduction} is
\begin{equation}
W_{\mathrm{iso}}(\boldsymbol{\epsilon})
=
\frac{1}{2}\boldsymbol{\sigma}:\boldsymbol{\epsilon}
=
\frac{\lambda}{2}
(\operatorname{tr}\boldsymbol{\epsilon})^2
+
\mu\,\boldsymbol{\epsilon}:\boldsymbol{\epsilon}.
\label{eq:isotropic_linear_energy}
\end{equation}
Requiring the two quadratic energies to agree for arbitrary symmetric small strains allows their coefficients to be matched independently, giving $\lambda=\mu=c/15$. Substituting this relation into the standard modulus conversions used in Eq.~\eqref{eq:cauchy_limit} recovers $\kappa=5\mu/3$ and $\nu=1/4$. Thus, the isotropic central-force network spans only a restricted one-parameter subset of isotropic linear elasticity: changing the spring stiffness scale changes the overall rigidity but not the relative volumetric and shear response. In particular, it cannot independently control $\kappa$ and $\mu$ within the unstressed, isotropic, affine small-strain regime considered here. With nonuniform edge weights, the effective response may become anisotropic, in which case it is no longer described by the isotropic Lamé parameters used above.

\subsection{Relaxed Energy of Response Carriers}
\label{app:carrier_energy}
The affine central-force restriction derived in
Sec.~\ref{app:central_coupling} arises when every node follows the
prescribed affine deformation. MorphTwin enriches
the admissible deformation modes by allowing
auxiliary carrier nodes to relax relative to 
original physical nodes. This internal freedom
changes how spring network stores energy
under different deformation modes.

We now derive the relaxed elastic energy in Eq.~\eqref{eq:relaxed_energy}. Let $\mathbf E$ denote a prescribed symmetric
small strain applied affinely to the original
physical nodes. It is kept generic in the derivation, while the volumetric probe $\mathbf E_{\mathrm{vol}}=\mathbf I/\sqrt{3}$ and five orthonormal trace-free probes $\{\mathbf E_{\mathrm{dev}}^{(a)}\}_{a=1}^{5}$ used in Eq.~\eqref{eq:effective_response} are particular choices of $\mathbf E$. For a spring $e=(i,j)\in\mathcal S$, let its reference edge vector be $\mathbf d_e=\mathbf x_j-\mathbf x_i=L_e\mathbf n_e$. In addition to affine displacement induced by $\mathbf E$, relaxable auxiliary nodes may undergo internal displacements $\mathbf z_i$, whereas the original physical nodes have $\mathbf z_i=\mathbf 0$. 

The first-order change of the edge vector is $\Delta\mathbf d_e=\mathbf E\mathbf d_e+(\mathbf z_j-\mathbf z_i)$, giving the first-order axial strain
\begin{equation}
\frac{\Delta L_e}{L_e}
=
\mathbf n_e^\top\mathbf E\mathbf n_e
+
\frac{\mathbf n_e^\top(\mathbf z_j-\mathbf z_i)}{L_e}.
\label{eq:carrier_axial_strain}
\end{equation}
Here, $\Delta L_e$ denotes the first-order change
in the length of edge $e$.
For a fixed internal relaxation $\mathbf z$, the corresponding quadratic central-spring energy is
\begin{equation}
\begin{aligned}
\mathcal E(\mathbf E,\mathbf z;\mathbf b,\mathbf u)
&=
\frac{1}{2}
\sum_{e=(i,j)\in\mathcal S}
Y_e(\mathbf b,\mathbf u)L_e
\\
&\quad \times
\left[
\mathbf n_e^\top\mathbf E\mathbf n_e
+
\frac{\mathbf n_e^\top(\mathbf z_j-\mathbf z_i)}{L_e}
\right]^2.
\end{aligned}
\label{eq:carrier_energy_unrelaxed}
\end{equation}
The physical nodes are constrained to the prescribed affine deformation, while the auxiliary nodes are free to relax. Eliminating these internal degrees of freedom gives the relaxed energy $\overline{\mathcal E}(\mathbf E;\mathbf b,\mathbf u)=\min_{\mathbf z} \mathcal E(\mathbf E,\mathbf z;\mathbf b,\mathbf u)$, as defined in Eq.~\eqref{eq:relaxed_energy}. Evaluating this same function at $\mathbf E_{\mathrm{vol}}$ and $\mathbf E_{\mathrm{dev}}^{(a)}$ yields the volumetric and deviatoric energies used to define $\kappa(\mathbf b,\mathbf u)$ and $\mu(\mathbf b,\mathbf u)$ in Eq.~\eqref{eq:effective_response}.

To make explicit how auxiliary relaxation modifies the response
to a prescribed strain probe, we rewrite the spring energy in
matrix form and eliminate the internal degrees of freedom.
Let $N_a$ denote the number of auxiliary nodes and
$m=|\mathcal S|$ the number of springs.
We stack the auxiliary displacements into
$\mathbf z\in\mathbb R^{3N_a}$, with zero displacement
assigned to the original physical nodes.
The matrix $\mathbf B\in\mathbb R^{m\times 3N_a}$
maps these displacements to the relaxation-induced
axial strains of individual springs, such that
$(\mathbf B\mathbf z)_e
=\mathbf n_e^\top(\mathbf z_j-\mathbf z_i)/L_e$
for each spring $e=(i,j)$.
Let $\mathbf g(\mathbf E)$ collect the affine axial strains
$g_e(\mathbf E)=\mathbf n_e^\top\mathbf E\mathbf n_e$,
and define
$K(\mathbf b,\mathbf u)
=
\operatorname{diag}(Y_eL_e)
=
\operatorname{diag}(k_eL_e^2)$.
The unrelaxed energy can then be written as
\begin{equation}
\mathcal E(\mathbf E,\mathbf z;\mathbf b,\mathbf u)
=
\frac{1}{2}
\left[\mathbf g(\mathbf E)+\mathbf B\mathbf z\right]^\top
\mathbf K(\mathbf b,\mathbf u)
\left[\mathbf g(\mathbf E)+\mathbf B\mathbf z\right].
\label{eq:carrier_energy_matrix}
\end{equation}
Minimizing the quadratic form over the auxiliary displacements $\mathbf z$ eliminates the internally relaxable degrees of freedom, yielding the reduced energy of the prescribed physical-node deformation:
\begin{equation}
\overline{\mathcal E}(\mathbf E;\mathbf b,\mathbf u)
=
\frac{1}{2}
\mathbf g(\mathbf E)^\top
\left[
\mathbf K
-
\mathbf K\mathbf B
\left(\mathbf B^\top\mathbf K\mathbf B\right)^\dagger
\mathbf B^\top\mathbf K
\right]
\mathbf g(\mathbf E),
\label{eq:carrier_energy_eliminated}
\end{equation}
where $\mathbf K=\mathbf K(\mathbf b,\mathbf u)$ for compactness and $(\cdot)^\dagger$ denotes the pseudoinverse when internal null modes are present. The bracketed operator is the effective stiffness acting on the affine edge-strain vector after the auxiliary degrees of freedom have relaxed. It separates the affine stiffness
from the reduction enabled by internal relaxation.
For the same augmented graph, let
$\mathcal E_{\rm aff}(\mathbf E)
=\frac12\mathbf g(\mathbf E)^\top
\mathbf K\mathbf g(\mathbf E)$
denote the energy with all nodes deforming affinely.
The difference between this energy and the
relaxed energy, which is derived as 
$
\mathcal E_{\rm aff}(\mathbf E)
-\overline{\mathcal E}(\mathbf E)
=\frac12
\big(\mathbf B^\top \mathbf K\mathbf g(\mathbf E)\big)^\top
(\mathbf B^\top \mathbf K \mathbf B)^\dagger
\big(\mathbf B^\top \mathbf K\mathbf g(\mathbf E)\big)
\geq 0$, is nonnegative and depends
on both the prescribed strain through
$\mathbf g(\mathbf E)$ and the carrier geometry
through $\mathbf B$.

Consequently, volumetric and deviatoric strains
can experience different energy reductions,
changing their effective response ratio beyond
the affine central-force restriction.
The two carrier families introduce distinct
relaxation patterns, whose local independence
is determined by the rank of the response
Jacobian in Eq.~\eqref{eq:response_calibration}.

\subsection{Volumetric and Deviatoric Response Modes}
\label{app:response_modes}

A symmetric $3\times3$ strain tensor has six independent components,
which decompose orthogonally into one volumetric direction and a
five-dimensional trace-free deviatoric subspace.
Since $\|\mathbf I\|_F=\sqrt{3}$, the unit volumetric probe in
Eq.~\eqref{eq:effective_response} is
$\mathbf E_{\mathrm{vol}}=\frac{\mathbf I}{\sqrt{3}}$.
For deviatoric subspace, we use orthonormal basis
\begin{equation}
\begin{aligned}
\mathbf E_{\mathrm{dev}}^{(1)}&=\frac{1}{\sqrt2}
\begin{bmatrix}
1&0&0\\
0&-1&0\\
0&0&0
\end{bmatrix},
&
\mathbf E_{\mathrm{dev}}^{(2)}&=\frac{1}{\sqrt6}
\begin{bmatrix}
1&0&0\\
0&1&0\\
0&0&-2
\end{bmatrix},
\\
\mathbf E_{\mathrm{dev}}^{(3)}&=\frac{1}{\sqrt2}
\begin{bmatrix}
0&1&0\\
1&0&0\\
0&0&0
\end{bmatrix},
&
\mathbf E_{\mathrm{dev}}^{(4)}&=\frac{1}{\sqrt2}
\begin{bmatrix}
0&0&1\\
0&0&0\\
1&0&0
\end{bmatrix},
\\
\mathbf E_{\mathrm{dev}}^{(5)}&=\frac{1}{\sqrt2}
\begin{bmatrix}
0&0&0\\
0&0&1\\
0&1&0
\end{bmatrix}.
\end{aligned}
\end{equation}
These probes satisfy
$\operatorname{tr}\mathbf E_{\mathrm{dev}}^{(a)}=0$ and
$\mathbf E_{\mathrm{dev}}^{(a)}:\mathbf E_{\mathrm{dev}}^{(b)}
=\delta_{ab}$.
For a general strain, define
$\mathbf E_{\mathrm{dev}}
=\mathbf E-\frac{1}{3}(\operatorname{tr}\mathbf E)\mathbf I$.
The corresponding isotropic strain-energy density is
\begin{equation}
W(\mathbf E)
=
\frac{\kappa}{2V}
\bigl(\operatorname{tr}\mathbf E\bigr)^2
+
\frac{\mu}{V}
\mathbf E_{\mathrm{dev}}:\mathbf E_{\mathrm{dev}},
\end{equation}
where $\kappa/V$ and $\mu/V$ are the corresponding continuum bulk and
shear moduli. The total elastic energy of the reference body is therefore
$
\mathcal E(\mathbf E)
=
V W(\mathbf E)
=
\frac{\kappa}{2}
\bigl(\operatorname{tr}\mathbf E\bigr)^2
+
\mu\,
\mathbf E_{\mathrm{dev}}:\mathbf E_{\mathrm{dev}}$. For the unit volumetric probe, $\operatorname{tr}(\mathbf E_{\mathrm{vol}})=\sqrt{3}$ and $\mathbf E_{\mathrm{dev}}=\mathbf 0$, giving $\mathcal E(\mathbf E_{\mathrm{vol}})=\frac{3}{2}\kappa$. Each unit deviatoric probe is trace-free and has unit Frobenius norm, giving $\mathcal E(\mathbf E_{\mathrm{dev}}^{(a)})=\mu$. These identities give exactly the normalization factors $2/3$ and $1/5$ used in Eq.~\eqref{eq:effective_response}: the volumetric response is recovered from the relaxed total energy under the volumetric probe, while the shear response is obtained by averaging the relaxed total energies over the five deviatoric probes.

For a perfectly isotropic response, all five deviatoric probe energies are
identical. For a discrete graph with residual anisotropy, averaging over any
complete orthonormal basis of the five-dimensional deviatoric subspace gives
a basis-independent scalar summary of its quadratic response.
Although absolute graph-integrated responses contain the reference-volume
factor, it is fixed for each graph and cancels in the relative logarithmic
change $\Delta\mathbf q=\mathbf q-\mathbf q^0$.
Thus, $\Delta\mathbf q$ measures the relative change in the corresponding
continuum bulk and shear moduli from the graph's calibration reference and
serves as the shared quantity in Sec.~\ref{sec:material_identification}.

\paragraph{Local response controllability and transfer.}
For a reconstructed graph $g$, let
$\mathbf q_g(\mathbf u)$ denote its logarithmic
bulk and shear responses with backbone stiffnesses
fixed at $\mathbf b_g^0$.
Near the reference state,
\begin{equation}
\mathbf q_g(\mathbf u)-\mathbf q_g^0
=
\mathbf A_g\mathbf u+O(\|\mathbf u\|^2),
\quad
\mathbf A_g=
\left.
\frac{\partial\mathbf q_g}{\partial\mathbf u}
\right|_{\mathbf u=\mathbf0}.
\end{equation}
Each column of $\mathbf A_g$ measures how one
carrier family affects both material responses.
When the columns are linearly independent,
the response mapping is locally invertible,
allowing independent bulk and shear control
through $\mathbf u_g\simeq
\mathbf A_g^{-1}\Delta\mathbf q_g$.

This mapping also enables response control
across different graphs. A shared relative edit
uses the same $\Delta\mathbf q$ with each graph's
own Jacobian, while matching an absolute response
target $\mathbf q^\star$ requires
$\Delta\mathbf q_g=\mathbf q^\star-\mathbf q_g^0$.
Finite-amplitude response matching uses the
nonlinear calibration described in Sec.~\ref{app:material_reuse_fracture}.

\section{Implementation Details}
\label{app:implementation}

This section details carrier construction and deployment,
material fitting, and Gaussian appearance binding.
The material-response formulation and graph-specific
calibration are introduced in
Secs.~\ref{sec:material_response}--\ref{sec:material_identification}.

\subsection{Carrier Construction and Deployment}
\label{app:carrier_construction}

Starting from the reconstructed graph
$\mathcal G_0=(\mathcal P_0,\mathcal S_0)$, we construct bulk and shear
carriers on local four-node cells.
Each valid cell generates a centroid bulk carrier and
candidate symmetric shear carriers.
Their relaxed bulk and shear responses are evaluated
using the static condensation in Sec.~\ref{app:carrier_energy}.
We retain geometrically valid groups and select shear
orientations with complementary deviatoric responses.

The retained groups are assigned family-wise reference
stiffness scales and stability-based auxiliary masses.
Groups without feasible mass ranges are removed.
We then jointly select and scale the carrier groups
while determining the mass outflow from each original
physical node.
The linear program minimizes total auxiliary mass subject
to the prescribed bulk and shear response contributions,
mass and first-moment conservation, inertia bounds,
and spring and deployment constraints.
Its solution is further checked using the assembled
mass, stiffness, and damping matrices.
If the spectral stability bound is violated, a modal
constraint is added and the program is solved again.

The selected group $g$ receives a total mass
$m_g=\alpha_g\overline m_g$, distributed equally among
its $n_g$ auxiliary nodes.
The LP also specifies the total mass $\ell_i$ transferred
from each physical node.
We realize these mass marginals using a deterministic
distance-prioritized greedy assignment, expanding the
candidate neighborhood when necessary.
Writing $t_{i,a}$ for the mass transferred from physical
node $i$ to auxiliary node $a$, the resulting allocation
satisfies
\begin{equation}
\sum_a t_{i,a}=\ell_i,\quad
\sum_i t_{i,a}=m_a,\quad
m_i=m_i^0-\ell_i.
\label{eq:carrier_mass_transfer}
\end{equation}
The LP preserves the reference first mass moment,
while the greedy assignment realizes the prescribed
mass marginals.
The original physical nodes and backbone connections
are retained, and the backbone and carrier coefficients
are jointly normalized to preserve the reference bulk
and shear responses.

Finally, each spring is assigned its family label
$\mathbf w_e$ and reference coefficient
$Y_e(\mathbf b,\mathbf 0)$ as defined in Eq.~\eqref{eq:carrier_materialization}.
The construction-time group scales determine the
deployed reference graph, whereas the material controls
$\mathbf u=(u_B,u_S)^\top$ subsequently edit the two
carrier families through Eq.~\eqref{eq:carrier_materialization}.
The resulting graph is simulated explicitly with both
original and auxiliary nodes.

\begin{algorithm}[t]
\caption{Carrier construction and deployment.}
\label{alg:carrier_construction}
\begin{algorithmic}[1]

\Require $\mathcal G_0=(\mathcal P_0,\mathcal S_0)$,
$\mathbf x^0$, $m_i^0$, and $\Delta t$

\State Generate local four-node cells and construct bulk
and shear carrier candidates
\State Evaluate relaxed responses; screen invalid groups
and select complementary shear orientations

\Repeat
    \State Assign reference stiffness scales and compute
    stability-based masses $\overline m_g$
    \State Remove groups with infeasible mass ranges
\Until{the candidate set stabilizes}

\State $(\boldsymbol\alpha,\boldsymbol\ell)
\gets \operatorname{SolveLP}$

\If{the LP is infeasible}
    \State Reject the construction
\EndIf

\While{the spectral stability bound is violated}
    \State Add the violated modal constraint and re-solve
    the LP
    \If{the LP is infeasible}
        \State Reject the construction
    \EndIf
\EndWhile

\ForAll{selected groups $g$}
    \State $m_g\gets\alpha_g\overline m_g$
    \State Distribute $m_g$ equally among its $n_g$
    auxiliary nodes
\EndFor

\State $\{t_{i,a}\}\gets
\operatorname{GreedyMassAssignment}(\boldsymbol\ell,\{m_a\})$
\State $m_i\gets m_i^0-\ell_i$
\State Normalize backbone and carrier coefficients
\State Assign $\mathbf w_e$ and $Y_e(\mathbf b,\mathbf 0)$
\State Verify mass, first-moment, and response-rank constraints

\Ensure Augmented graph $\mathcal G$

\end{algorithmic}
\end{algorithm}
\subsection{Material Fitting}
\label{app:material_fitting}

\paragraph{Initialization and graph calibration.}
Covariance matrix adaptation evolution strategy (CMA-ES) selects
the initial spring scale and contact settings by minimizing rollout
error on the modeling frames. At the reference state
$(\mathbf b^0,\mathbf 0)$, we compute
$\mathbf q^0=\mathbf q(\mathbf b^0,\mathbf 0)$ and the response
Jacobian $\mathbf A$ as in Sec.~\ref{sec:material_identification}.
The local map $\mathbf u=\mathbf A^{-1}\Delta\mathbf q$ converts
requested log changes in bulk and shear response into carrier controls.
We fix $\mathbf q^0$, $\mathbf A$, carrier geometry, and edge-family
assignments throughout fitting.

\paragraph{Physics optimization.}
Starting from $\mathbf b^0$ and $\Delta\mathbf q=\mathbf 0$,
Adam jointly fits $\Delta\mathbf q$, backbone stiffnesses $\mathbf b$,
and interaction parameters $\boldsymbol\eta$.
These interaction parameters include one stiffness per spring linking
the hand to the object, plus the learned friction and restitution
coefficients for ground contact and self-contact.
Friction lies in $[0,2]$ and restitution in $[0,1]$.
Both are dimensionless. Hand trajectories are prescribed,
and masses, damping, and contact distance remain fixed.

% At each step, Eq.~3 in the main paper gives the spring coefficients
% $\mathbf Y(\mathbf b,\mathbf u)$. Explicit simulation produces
% node trajectories $\widehat{\mathbf X}$, which we fit to the
% reconstructed geometry and 3D tracks using
% $\mathcal L_{\mathrm{motion}}=\mathcal L_{\mathrm{geo}}+\mathcal L_{\mathrm{track}}$.
% As $\mathbf b$ changes, the realized response is
% $\mathbf q(\mathbf b,\mathbf u)$.
\paragraph{Explicit spring-mass simulation.}
For each forward pass, Eq.~\eqref{eq:carrier_materialization}
determines the spring stiffnesses $\mathbf Y(\mathbf b,\mathbf u)$.
All physical and auxiliary nodes follow the same
central-force dynamics:
\begin{equation}
\begin{aligned}
m_i\ddot{\mathbf x}_i^t
={}&
\mathbf f_i^{\mathrm{spr}}(\mathbf X^t;\mathbf Y)
+\mathbf f_i^{\mathrm{damp}}(\mathbf X^t,\dot{\mathbf X}^t)\\
&+\mathbf f_i^{\mathrm{int}}
(\mathbf X^t,\dot{\mathbf X}^t;\boldsymbol\eta)
+m_i\mathbf g,
\end{aligned}
\label{eq:explicit_dynamics}
\end{equation}
where $\mathbf X^t$ stacks all node positions,
$m_i$ is the node mass, and $\mathbf g$ is gravity.
The three force terms denote elastic, damping,
and interaction forces, respectively. Combining these forces with gravity gives the total force $\mathbf f_i^t$, which is used to advance the node states through semi-implicit Euler integration:
\begin{equation}
\mathbf v_i^{t+\Delta t}
=\mathbf v_i^t
+\frac{\Delta t}{m_i}\mathbf f_i^t,\quad
\mathbf x_i^{t+\Delta t}
=\mathbf x_i^t
+\Delta t\,\mathbf v_i^{t+\Delta t},
\label{eq:explicit_integration}
\end{equation}
where $\mathbf f_i^t$ is the total force in
Eq.~\eqref{eq:explicit_dynamics}, including gravity.
Each frame transition uses 667 substeps with
$\Delta t=5\times10^{-5}$~s, as listed in Table~\ref{tab:optimization_settings}.
The resulting differentiable trajectory
$\widehat{\mathbf X}(\boldsymbol\theta)
=\operatorname{Simulate}(\mathcal G,\mathbf Y,\boldsymbol\eta)$
is fitted to reconstructed geometry and 3D tracks
using $\mathcal L_{\mathrm{motion}}
=\mathcal L_{\mathrm{geo}}+\mathcal L_{\mathrm{track}}$.
As $\mathbf b$ changes, the realized material
response is $\mathbf q(\mathbf b,\mathbf u)$.

\paragraph{Appearance fitting.}
We freeze the fitted physics and optimize the canonical Gaussian
appearance parameters $\boldsymbol\varphi$. The binding in
Sec.~\ref{app:gaussian_binding} deforms the Gaussians along the
simulated trajectories. For each training frame and view,
$\widehat{\mathbf I}$ denotes the rendered image and $\mathbf I$
the observed RGB image masked to the object. Hand pixels are
zeroed in both images. Following 3D Gaussian Splatting~\citep{kerbl20233dgs},
we use
\begin{equation}
\mathcal L_{\mathrm{RGB}}
= 0.8\,\mathcal L_1(\widehat{\mathbf I},\mathbf I)
+ 0.2\,\mathcal L_{\mathrm{SSIM}}(\widehat{\mathbf I},\mathbf I).
\end{equation}

\paragraph{Inference and material morphing.}
Future prediction uses the fitted physics directly.
For local edits, each graph maps the requested response change
through its own fixed Jacobian. For finite-amplitude transfer,
we measure the fitted source response and match it on the target
graph using the offline calibration in
Sec.~\ref{app:material_reuse_fracture}.

\subsection{Optimization Settings}
\label{app:optimization_settings}

Table~\ref{tab:optimization_settings} summarizes the numerical settings used
for material fitting. Dataset-specific appearance-training and evaluation
settings are reported separately in Sec.~\ref{app:experiments}.

\begin{table}[htbp]
\centering
\small
\caption{
Optimization settings for MorphTwin training.
}
\label{tab:optimization_settings}

\setlength{\tabcolsep}{4pt}
\renewcommand{\arraystretch}{1.08}

\begin{tabular*}{\columnwidth}{
@{\extracolsep{\fill}}lr@{}
}
\toprule
Setting & Value \\
\midrule

Optimizer & Adam \\
Material-fitting epochs & $188$ \\
Shared-response learning rate & $10^{-3}$ \\
Local-parameter learning rate & $10^{-3}$ \\
Adam $(\beta_1,\beta_2)$ & $(0.9,\,0.99)$ \\
Adam $\epsilon$ & $10^{-8}$ \\
Video rate & $30\,\mathrm{Hz}$ \\
Simulation substep $\Delta t$ & $5\times10^{-5}\,\mathrm{s}$ \\
Learning-rate decay & None \\
Substeps per frame transition & $667$ \\
\bottomrule
\end{tabular*}
\end{table}

\subsection{Optimization and Inference Procedure}
\label{app:optimization_algorithm}
Algorithm~\ref{alg:morphtwin_optimization}
summarizes the material fitting, appearance optimization,
and subsequent inference procedures.
Graph calibration is performed once before fitting,
and the calibrated graph is reused for prediction
and material editing.

\begin{algorithm}[tbp]
\caption{MorphTwin optimization and inference.}
\label{alg:morphtwin_optimization}
\begin{algorithmic}[1]

\Require Observations $\mathcal V$ and initial graph
$\mathcal G_0=(\mathcal P_0,\mathcal S_0)$

\Statex \textbf{Graph and carrier initialization}
\State Initialize $\mathbf b^0$ and $\boldsymbol\eta$
from Sec.~\ref{sec:physics_twin}
\State $\mathcal G\gets\operatorname{Augment}(\mathcal G_0)$

\Statex \textbf{Energy response calibration}
\State $\mathbf q^0\gets\mathbf q(\mathbf b^0,\mathbf 0)$, $\mathbf A\gets
\left.\frac{\partial\mathbf q}{\partial\mathbf u}
\right|_{(\mathbf b^0,\mathbf 0)}$
\State Fix $\mathbf q^0$, $\mathbf A$, and carrier geometry

\Statex \textbf{Physics fitting}
\State $\mathbf b\gets\mathbf b^0$,
$\Delta\mathbf q\gets\mathbf 0$
\Repeat
    \State $\mathbf u\gets\mathbf A^{-1}\Delta\mathbf q$
    \State $\mathbf Y\gets
        \operatorname{Materialize}(\mathbf b,\mathbf u)$
    \State $\widehat{\mathbf X}\gets
        \operatorname{Simulate}
        (\mathcal G,\mathbf Y,\boldsymbol\eta)$
    \State Compute $\mathcal L_{\rm geo}$ and
        $\mathcal L_{\rm track}$
    \State Update $\Delta\mathbf q,\mathbf b,\boldsymbol\eta$
        using $\mathcal L_{\rm motion}$
\Until{convergence}

\Statex \textbf{Appearance fitting}
\State Freeze $\boldsymbol\theta=
(\Delta\mathbf q,\mathbf b,\boldsymbol\eta)$
\State Bind canonical Gaussians to $\mathcal P_0$
\State Optimize $\boldsymbol\varphi$ using
$\mathcal L_{\rm RGB}$ with fixed physics

\Statex \textbf{Prediction, editing, and transfer}
\If{controlled response transfer}
    \State Set target from the source material response
    \State Obtain $\mathbf Y'$ using the target graph's
    offline response table
\Else
    \State Choose $\Delta\mathbf q'$
    \Comment{$\Delta\mathbf q$ for prediction}
    \State $\mathbf u'\gets\mathbf A^{-1}\Delta\mathbf q'$
    \State $\mathbf Y'\gets
        \operatorname{Materialize}(\mathbf b,\mathbf u')$
\EndIf
\State $\widehat{\mathbf X}'\gets
    \operatorname{Simulate}
    (\mathcal G,\mathbf Y',\boldsymbol\eta)$

\end{algorithmic}
\end{algorithm}

\subsection{Gaussian Deformation and Binding}
\label{app:gaussian_binding}
For our experiments, Gaussian appearance is driven by the
trajectories of the original physical nodes $\mathcal P_0$.
Auxiliary carrier nodes participate in dynamics but are not
used as appearance-binding supports.
For each reference Gaussian center $\mathbf g_j^0$, we fix
its $K=16$ nearest physical nodes in the reference configuration,
denoted by $\mathcal N_j$, and assign normalized inverse-distance
weights $w_{ji}\propto
(\|\mathbf g_j^0-\mathbf x_i^0\|_2+10^{-6})^{-1}$,
with $\sum_{i\in\mathcal N_j}w_{ji}=1$.
The neighborhood and weights remain fixed over time.
We compute the reference and current weighted centroids
$\overline{\mathbf x}_j^0=\sum_{i\in\mathcal N_j}w_{ji}\mathbf x_i^0$
and
$\overline{\mathbf x}_j^t=\sum_{i\in\mathcal N_j}w_{ji}\mathbf x_i^t$,
and denote the centered positions by
$\mathbf r_{ji}=\mathbf x_i^0-\overline{\mathbf x}_j^0$
and
$\mathbf y_{ji}^t=\mathbf x_i^t-\overline{\mathbf x}_j^t$.
The weighted reference covariance and cross-covariance are:
\begin{equation}
\mathbf C_j=
\sum_{i\in\mathcal N_j}
w_{ji}\mathbf r_{ji}\mathbf r_{ji}^{\top},
\quad
\mathbf H_j^t=
\sum_{i\in\mathcal N_j}
w_{ji}\mathbf y_{ji}^t\mathbf r_{ji}^{\top}.
\end{equation}

We estimate the local rotation $\mathbf R_j^t$ using
weighted orthogonal Procrustes alignment.
Specifically, if
$\mathbf H_j^t=\mathbf U\mathbf\Sigma\mathbf V^\top$
is its singular value decomposition, the proper rotation is
\begin{equation}
\mathbf R_j^t
=
\mathbf U
\operatorname{diag}
\left(1,1,\det(\mathbf U\mathbf V^\top)\right)
\mathbf V^\top.
\end{equation}

The local affine map is estimated by weighted
least-squares fitting.
Since the reference neighborhood may not span all
three spatial directions, we use a truncated
pseudoinverse $\mathbf C_j^+$, which inverts the
covariance only along sufficiently supported directions.
The complementary directions are supplied by the
local rotation $\mathbf R_j^t$.
The resulting deformation gradient is
\begin{equation}
\begin{aligned}
\mathbf F_j^t
&=
\mathbf H_j^t\mathbf C_j^+
+
\mathbf R_j^t
\left(
\mathbf I-\mathbf C_j\mathbf C_j^+
\right),\\
\mathbf g_j^t
&=
\overline{\mathbf x}_j^t
+
\mathbf F_j^t
\left(
\mathbf g_j^0+\boldsymbol\delta_j
-\overline{\mathbf x}_j^0
\right),\\
\mathbf\Sigma_j^t
&=
\mathbf F_j^t
\mathbf\Sigma_j^0
(\mathbf F_j^t)^\top,
\end{aligned}
\end{equation}
where $\boldsymbol\delta_j$ is an optional learned
reference-space offset that adjusts Gaussian's
binding location and is shared across frames.
The covariance transformation transfers local stretching,
compression, and shear to Gaussian shape and orientation.
Each frame is computed directly from the reference state. Thus, the deformation produces updated Gaussian centers $\mathbf g_j^t$
and covariances $\mathbf\Sigma_j^t$ at each frame $t$.
Together with their learned appearance attributes,
these Gaussians are passed to the renderer to synthesize
corresponding image.

PhysTwin instead blends local rigid transformations and
updates Gaussian orientation through blended rotations,
without directly transporting covariance using a local
affine deformation gradient.
Our formulation applies $\mathbf F_j^t$ to both Gaussian
centers and covariances, allowing local affine deformation
beyond rigid motion.
The two pipelines therefore share the same physics-then-
appearance optimization strategy but differ in their
appearance-deformation models.

\section{Dataset Details}
\label{app:dataset}
We introduce a self-captured egocentric interaction
benchmark comprising 15 sequences recorded with
egocentric stereo RGB and an external RGB-D camera.
The observations are temporally aligned and
registered into consistent spatial coordinate
systems through post processing, without assuming
hardware-level synchronization.

\paragraph{Capture setup and raw recording.}
We build a first-person interaction capture system centered on a Meta Quest 3 headset and a data-processing workstation, together with an external Orbbec DaBai RGB-D camera that provides the main third-person RGB-D view. On the headset, stereo RGB images are acquired through the camera interface, while the device depth and motion-tracking interfaces record depth maps, head pose, articulated hand joints, and body pose. We additionally store camera intrinsics and extrinsics together with timestamps for all recorded streams. During capture, an operator manipulates deformable objects on a tabletop experimental platform equipped with a robotic arm through grasping, squeezing, and stretching interactions, and the acquisition program packages the image and tracking streams into multi-track recording files. The raw egocentric RGB streams are recorded at 60 fps with a per-eye resolution of $1280\times960$, while headset depth is recorded at $320\times320$ with an effective sampling rate of approximately 4.9 Hz in the current collection.

\paragraph{Post-processing and aligned observations.}
During post-processing, we first retain raw recordings with a duration of at least 20 s. We then remove non-interaction segments before and after the manipulation interval and extract the resulting interaction sequences for subsequent processing and evaluation. Subsequently, we decode the recorded streams, match observations by timestamp, verify pose validity, and transform all measurements into consistent camera and world coordinate systems. Depth observations are back-projected into 3D and registered to the corresponding RGB images, producing colored point clouds in both the world coordinate frame and the egocentric camera frame for subsequent object segmentation, deformation tracking, and physical modeling. The Meta Quest 3 streams and the main DaBai RGB-D stream are temporally aligned in post-processing, although hardware-level synchronization is not assumed. For the final experiments, all sequences are processed at $640\times360$ with a nominal frame rate of 30 fps. AI agents assist with keypoint annotation and temporal alignment, with every processing stage undergoing systematic manual review and correction where necessary. The resulting aligned multi-view observations support physical twin reconstruction and future prediction evaluation using the egocentric views, the RGB-D view, or their combination.

\paragraph{Benchmark composition and visualization.}
\begin{table}[t]
\centering
\caption{Overview of the 15 interaction sequences in our egocentric interaction benchmark.}
\label{tab:dataset-overview}

\small
\setlength{\tabcolsep}{3pt}
\renewcommand{\arraystretch}{1.12}

\resizebox{\columnwidth}{!}{%
\begin{tabular}{@{}lllS[table-format=4.0]S[table-format=3.2]@{}}
\toprule
Object & Morphology & Manipulation & {Frames} & {Duration (s)} \\
\midrule
Fabric Sheet (Trial 1) & Thin-shell & Bimanual & 196 & 6.53 \\
Fabric Sheet (Trial 2) & Thin-shell & Bimanual & 324 & 10.80 \\
Black Fabric           & Thin-shell & Bimanual & 361 & 12.03 \\
Mesh Fabric            & Thin-shell & Bimanual & 496 & 16.53 \\
Cushion (Trial 1)      & Volumetric & Bimanual & 118 & 3.93 \\
Cushion (Trial 2)      & Volumetric & Bimanual & 521 & 17.37 \\
Neck Pillow            & Volumetric & Bimanual & 471 & 15.70 \\
Plush Toy (Trial 1)    & Volumetric & Unimanual & 501 & 16.70 \\
Plush Toy (Trial 2)    & Volumetric & Bimanual & 251 & 8.37 \\
Mousepad                 & Volumetric & Bimanual & 431 & 14.37 \\
Book                   & Volumetric & Bimanual & 356 & 11.87 \\
Water Bottle           & Volumetric & Unimanual & 176 & 5.87 \\
Fabric Hat                & Thin-shell & Bimanual & 695 & 23.17 \\
Headphones             & Composite & Bimanual & 436 & 14.53 \\
Smartphone             & Volumetric & Unimanual & 407 & 13.57 \\
\midrule
\multicolumn{3}{@{}l}{Total: 15 sequences} & 5740 & 191.33 \\
\bottomrule
\end{tabular}%
}
\end{table}

\begin{figure*}[htbp]
\centering
\includegraphics[width=\textwidth]{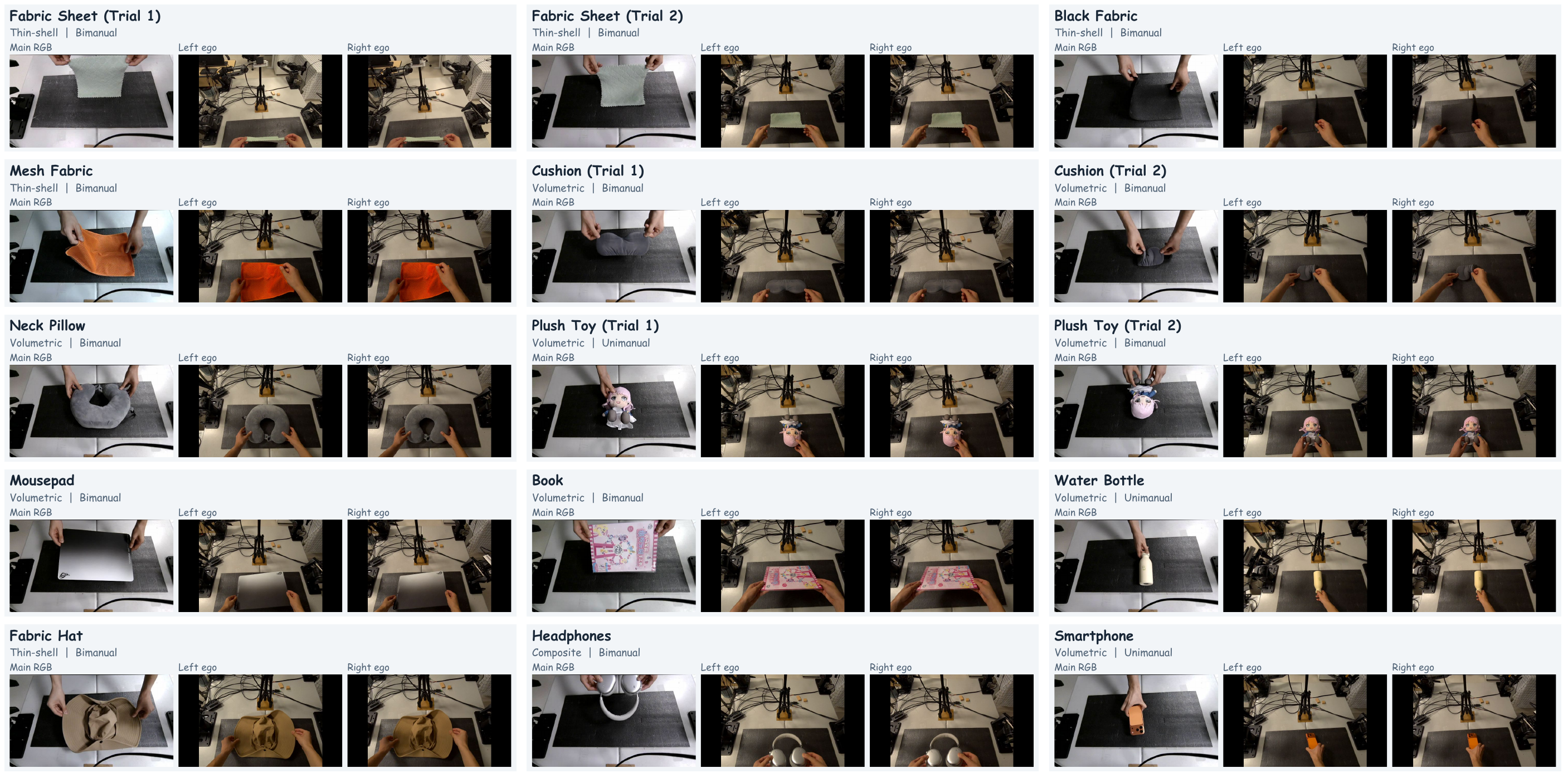}
\caption{
Visual overview of our 15 self-captured, spatiotemporally aligned egocentric RGB-D sequences listed in Table~\ref{tab:dataset-overview}.
For each sequence, we show a representative temporally aligned sample
from the main RGB view and the left and right egocentric views.
The morphology and manipulation labels shown in each panel follow the
terminology used in Table~\ref{tab:dataset-overview}.
}
\label{fig:dataset-overview}
\end{figure*}

Table~\ref{tab:dataset-overview} summarizes the 15 interaction sequences used in the dataset, including object morphology, manipulation type, number of frames, and sequence duration. Figure~\ref{fig:dataset-overview} provides a corresponding visual overview, showing one representative temporally aligned sample from the three camera views for each sequence.
Sequence duration is computed as the number of output frames divided by 30, with the total duration calculated before rounding. Frame counts refer to temporal samples within a sequence rather than the sum across camera views. Each temporal sample contains one main RGB-D observation from the DaBai camera together with the left and right egocentric RGB observations from the Meta Quest 3 headset. We use an approximately 70/30 temporal observation/future split.

\paragraph{Morphology and manipulation labels.}
The morphology labels describe coarse object form rather than measured physical properties. \emph{Thin-shell} denotes objects dominated by a thin deformable surface, \emph{volumetric} denotes objects with substantial three-dimensional bulk, and \emph{composite} denotes multi-part structures. Manipulation is categorized as \emph{unimanual} or \emph{bimanual} according to whether one or two hands are involved, while trial numbers distinguish separate recordings of the same object type.

\section{Experimental Protocol and Results}
\label{app:experiments}
\subsection{Experimental Protocol}
\paragraph{Real-video evaluation.}
Following DeformMaster~\citep{li2026deformmaster}, we select the same 20
PhysTwin interaction scenes to maintain a consistent scene set for comparison.
Let $T_{\mathrm{obs}}$ denote the observation boundary and $T$ the sequence
length. Reconstruction image metrics use frames $[0,T_{\mathrm{obs}})$,
whereas geometric and tracking metrics exclude the initial frame and use
$[1,T_{\mathrm{obs}})$. Future metrics use $[T_{\mathrm{obs}},T)$.

MorphTwin fits physics before optimizing Gaussian appearance with the
physical trajectories fixed. Future rollout uses prescribed interaction
controls without fitting to future object observations.

\paragraph{Synthetic material and appearance evaluation.}
We use 14 simulation-generated Spring-Gaus development scenes to evaluate
material recovery. Their known Young's moduli and Poisson's ratios provide
ground-truth references that are unavailable for our real-world recordings.
These references enable direct quantitative validation of material recovery
using errors in both parameters.
MorphTwin and the central-spring PhysTwin adaptation
share an inverse frontend supervised by observed particle trajectories,
with geometry, initial conditions, and the density assumption fixed.

As defined in Sec.~\ref{sec:material_identification} and Sec.~\ref{app:response_modes}, the recovered
$\kappa$ and $\mu$ are graph-integrated elastic responses, while
$\kappa/V$ and $\mu/V$ are the corresponding continuum moduli. Dynamic
observations constrain elastic stiffness relative to mass, so reporting an
absolute Young's modulus additionally requires a density convention. We use
the fixed density of $1000\,\mathrm{kg/m^3}$ for all synthetic material
recovery experiments. The reported absolute Young's-modulus values are
therefore conditional on this density assumption, whereas Poisson's ratio
and the relative logarithmic response change $\Delta\mathbf q$ are
independent of the common absolute stiffness-mass scale.

Methods share a canonical initialization fitted
for 30,000 iterations, followed by 10,000 appearance-fitting iterations per
method. Cameras other than camera 5 provide training images, while camera 5 is
held out for evaluation at $512\times512$. Frames 0--19 are observed and
frames 20--29 are future. We use the prespecified final appearance checkpoint.

\paragraph{Egocentric benchmark evaluation.}
The current collection contains 15 sequences and 5,740 temporal samples, corresponding to 191.33 seconds at a nominal 30 fps. Each sample includes left and right egocentric views and a main RGB-D view, processed at $640\times360$. The streams are temporally aligned, and hardware synchronization is not assumed. We use an approximately 70/30 temporal observation/future split, with frame counts referring to temporal samples. For each input views configuration, geometry reconstruction and the resulting 3D motion tracks are obtained by fusing only the designated input streams: the main RGB-D and two egocentric views for \emph{All Views}, the two egocentric views for \emph{Egocentric}, and the main RGB-D stream for \emph{Single RGB-D}. 

The same designated views supervise Gaussian appearance optimization, while PSNR, SSIM, and LPIPS are evaluated over all three views under a common protocol. Future rollout does not use future object geometry, motion, or appearance observations. It is conditioned on prescribed interaction controls calibrated from the complete interaction recording and then held fixed during prediction.

\paragraph{Metric definitions.}
Local PhysTwin evaluation uses camera 0, black backgrounds, hand exclusion,
and AlexNet LPIPS. Chamfer distance is the directed visible-ground-truth-to-
prediction nearest-neighbor $L_1$ distance in meters. Tracking error is the
mean $L_2$ distance using fixed particle associations established at frame 0.
Spring-Gaus appearance evaluation uses VGG LPIPS, and its separate geometry
diagnostics use symmetric unsquared Chamfer distance and correspondence
RMSE in millimeters. These geometry metrics are not interchangeable across
datasets. Material errors are
$100|\hat E-E_{\mathrm{GT}}|/E_{\mathrm{GT}}$ and
$|\hat\nu-\nu_{\mathrm{GT}}|$. Local aggregate scores weight scenes equally.
Controlled response and cross-graph experiments are reported separately
from video metrics, with their controls and error definitions specified
alongside the corresponding experiments.

\begingroup

% ---------------------------------------------------------
% Subtle separator between object groups
% ---------------------------------------------------------
\newcommand{\scenerule}{%
    \arrayrulecolor{black!24}%
    \hline
    \arrayrulecolor{black}%
}

\begin{table*}[t]
\centering
\scriptsize
\color{black}

% ---------------------------------------------------------
% Spacing
% ---------------------------------------------------------
\setlength{\tabcolsep}{3.6pt}
\renewcommand{\arraystretch}{1.04}

\setlength{\aboverulesep}{0.30ex}
\setlength{\belowrulesep}{0.30ex}
\setlength{\belowcaptionskip}{4pt}

\caption{
Quantitative evaluation on 15 sequences in our egocentric interaction benchmark under All Views supervision.
CD and Track are in meters.
We highlight the \textbf{best} and \underline{second-best} results for each metric and task.
}
\label{tab:morphtwin_per_scene_all_views}

\begin{adjustbox}{max width=\textwidth}

\begin{tabular}{
@{}
>{\raggedright\arraybackslash}p{1.65cm}
@{\hspace{1.5pt}}
>{\raggedright\arraybackslash}p{2.05cm}
@{\hspace{5pt}}
ccccc
@{\hspace{8pt}}
ccccc
@{}
}

\toprule

\rowcolor{gray!12}
& &
\multicolumn{5}{c}{\textbf{Reconstruction}}
& \multicolumn{5}{c}{\textbf{Future Prediction}} \\

\cmidrule(lr){3-7}
\cmidrule(lr){8-12}

Object & Method
& PSNR $\uparrow$
& SSIM $\uparrow$
& LPIPS $\downarrow$
& CD $\downarrow$
& Track $\downarrow$
& PSNR $\uparrow$
& SSIM $\uparrow$
& LPIPS $\downarrow$
& CD $\downarrow$
& Track $\downarrow$ \\

\midrule

% =========================================================
% Fabric Sheet (Trial 1)
% =========================================================
\cellcolor{gray!5}
& PhysTwin
& \underline{25.94} & \underline{0.9608} & \underline{0.0481}
& \underline{0.01002} & \underline{0.0231}
& \underline{20.85} & \underline{0.9366} & \underline{0.0994}
& \textbf{0.03182} & \underline{0.0629} \\

\cellcolor{gray!5}
& PGND
& 23.41 & 0.9563 & 0.0623
& 0.02534 & 0.0330
& 19.98 & 0.9327 & 0.1110
& 0.10498 & 0.0989 \\

\cellcolor{gray!5}
& GIC
& 21.51 & 0.9501 & 0.0952
& 0.04956 & 0.1125
& 19.97 & 0.9354 & 0.1143
& 0.17875 & 0.3177 \\

\rowcolor{blue!6}
\cellcolor{gray!5}
\multirow[c]{-4}{1.65cm}{%
\raggedright Fabric Sheet\\[-1pt](Trial 1)}
& \textbf{MorphTwin (Ours)}
& \textbf{31.72} & \textbf{0.9819} & \textbf{0.0390}
& \textbf{0.00948} & \textbf{0.0201}
& \textbf{22.54} & \textbf{0.9526} & \textbf{0.0805}
& \underline{0.03513} & \textbf{0.0626} \\

\scenerule

% =========================================================
% Fabric Sheet (Trial 2)
% =========================================================
\cellcolor{gray!5}
& PhysTwin
& \underline{30.28} & 0.9561 & \textbf{0.0350}
& \textbf{0.00490} & \textbf{0.0095}
& \underline{23.09} & \underline{0.9157} & \underline{0.0955}
& \underline{0.02071} & \underline{0.0441} \\

\cellcolor{gray!5}
& PGND
& 30.27 & \underline{0.9566} & 0.0384
& 0.00846 & \underline{0.0111}
& 19.89 & 0.8655 & 0.1889
& 0.12696 & 0.1446 \\

\cellcolor{gray!5}
& GIC
& 24.63 & 0.9378 & 0.1180
& 0.02361 & 0.0546
& 16.57 & 0.8595 & 0.2308
& 0.13755 & 0.2707 \\

\rowcolor{blue!6}
\cellcolor{gray!5}
\multirow[c]{-4}{1.65cm}{%
\raggedright Fabric Sheet\\[-1pt](Trial 2)}
& \textbf{MorphTwin (Ours)}
& \textbf{33.72} & \textbf{0.9778} & \underline{0.0362}
& \underline{0.00562} & 0.0119
& \textbf{24.92} & \textbf{0.9518} & \textbf{0.0681}
& \textbf{0.01806} & \textbf{0.0438} \\

\scenerule

% =========================================================
% Black Fabric
% =========================================================
\cellcolor{gray!5}
& PhysTwin
& \underline{33.31} & \underline{0.9300} & \textbf{0.0855}
& \underline{0.01221} & 0.0184
& \underline{31.96} & \underline{0.9260} & \underline{0.0971}
& \underline{0.02013} & \underline{0.0452} \\

\cellcolor{gray!5}
& PGND
& 32.67 & 0.9255 & 0.0963
& 0.01477 & \underline{0.0173}
& 29.81 & 0.8943 & 0.1658
& 0.04223 & 0.0675 \\

\cellcolor{gray!5}
& GIC
& 29.46 & 0.8886 & 0.1955
& 0.04795 & 0.0843
& 29.01 & 0.8832 & 0.1984
& 0.02902 & 0.1325 \\

\rowcolor{blue!6}
\cellcolor{gray!5}
\multirow[c]{-4}{1.65cm}{\raggedright Black Fabric}
& \textbf{MorphTwin (Ours)}
& \textbf{37.82} & \textbf{0.9624} & \underline{0.0935}
& \textbf{0.00947} & \textbf{0.0149}
& \textbf{35.70} & \textbf{0.9545} & \textbf{0.0872}
& \textbf{0.01272} & \textbf{0.0318} \\

\scenerule

% =========================================================
% Mesh Fabric
% =========================================================
\cellcolor{gray!5}
& PhysTwin
& 22.65 & 0.8521 & 0.1524
& \underline{0.01591} & \underline{0.0333}
& \underline{22.31} & 0.8513 & \textbf{0.1785}
& \textbf{0.01654} & \textbf{0.0360} \\

\cellcolor{gray!5}
& PGND
& \underline{24.74} & \underline{0.8677} & \underline{0.1303}
& 0.01654 & \textbf{0.0208}
& 21.77 & \underline{0.8552} & 0.2078
& 0.04064 & 0.0475 \\

\cellcolor{gray!5}
& GIC
& 16.97 & 0.8132 & 0.3455
& 0.06378 & 0.2361
& 16.57 & 0.8438 & 0.3445
& 0.10670 & 0.5027 \\

\rowcolor{blue!6}
\cellcolor{gray!5}
\multirow[c]{-4}{1.65cm}{\raggedright Mesh Fabric}
& \textbf{MorphTwin (Ours)}
& \textbf{28.53} & \textbf{0.9061} & \textbf{0.1231}
& \textbf{0.01574} & 0.0340
& \textbf{22.52} & \textbf{0.8679} & \underline{0.1939}
& \underline{0.01675} & \underline{0.0408} \\

\scenerule

% =========================================================
% Cushion (Trial 1)
% =========================================================
\cellcolor{gray!5}
& PhysTwin
& \underline{36.88} & \underline{0.9802} & \underline{0.0285}
& \underline{0.00492} & \textbf{0.0075}
& \textbf{31.95} & \underline{0.9622} & \underline{0.0468}
& \underline{0.01542} & \underline{0.0239} \\

\cellcolor{gray!5}
& PGND
& 33.68 & 0.9670 & 0.0478
& 0.01119 & 0.0205
& 25.04 & 0.8996 & 0.1558
& 0.16596 & 0.1679 \\

\cellcolor{gray!5}
& GIC
& 32.69 & 0.9606 & 0.0659
& 0.03583 & 0.0718
& 25.05 & 0.9321 & 0.1081
& 0.05579 & 0.1172 \\

\rowcolor{blue!6}
\cellcolor{gray!5}
\multirow[c]{-4}{1.65cm}{%
\raggedright Cushion\\[-1pt](Trial 1)}
& \textbf{MorphTwin (Ours)}
& \textbf{41.13} & \textbf{0.9900} & \textbf{0.0146}
& \textbf{0.00485} & \underline{0.0080}
& \underline{31.64} & \textbf{0.9689} & \textbf{0.0412}
& \textbf{0.01070} & \textbf{0.0190} \\

\scenerule

% =========================================================
% Cushion (Trial 2)
% =========================================================
\cellcolor{gray!5}
& PhysTwin
& \underline{35.71} & \underline{0.9802} & \underline{0.0261}
& \underline{0.00865} & \underline{0.0144}
& \underline{30.36} & 0.9602 & \underline{0.0692}
& \underline{0.01606} & \underline{0.0394} \\

\cellcolor{gray!5}
& PGND
& 34.30 & 0.9767 & 0.0380
& 0.01453 & 0.0147
& 30.00 & \textbf{0.9627} & 0.0777
& 0.05520 & 0.0455 \\

\cellcolor{gray!5}
& GIC
& 30.65 & 0.9607 & 0.0720
& 0.03134 & 0.0648
& 24.94 & 0.9354 & 0.1157
& 0.05321 & 0.1357 \\

\rowcolor{blue!6}
\cellcolor{gray!5}
\multirow[c]{-4}{1.65cm}{%
\raggedright Cushion\\[-1pt](Trial 2)}
& \textbf{MorphTwin (Ours)}
& \textbf{40.22} & \textbf{0.9893} & \textbf{0.0172}
& \textbf{0.00514} & \textbf{0.0101}
& \textbf{30.61} & \underline{0.9612} & \textbf{0.0673}
& \textbf{0.01281} & \textbf{0.0332} \\

\scenerule

% =========================================================
% Neck Pillow
% =========================================================
\cellcolor{gray!5}
& PhysTwin
& \underline{27.70} & \underline{0.9390} & \underline{0.0620}
& \underline{0.00945} & \textbf{0.0157}
& \underline{24.97} & \underline{0.9181} & \underline{0.1018}
& \textbf{0.01329} & \textbf{0.0253} \\

\cellcolor{gray!5}
& PGND
& 23.91 & 0.9078 & 0.1278
& 0.02112 & 0.0282
& 20.66 & 0.8590 & 0.2228
& 0.04803 & 0.0762 \\

\cellcolor{gray!5}
& GIC
& 24.26 & 0.9196 & 0.1173
& 0.02885 & 0.0553
& 23.47 & 0.9078 & 0.1402
& 0.03271 & 0.0677 \\

\rowcolor{blue!6}
\cellcolor{gray!5}
\multirow[c]{-4}{1.65cm}{\raggedright Neck Pillow}
& \textbf{MorphTwin (Ours)}
& \textbf{29.72} & \textbf{0.9605} & \textbf{0.0532}
& \textbf{0.00925} & \underline{0.0178}
& \textbf{25.46} & \textbf{0.9317} & \textbf{0.0915}
& \underline{0.01366} & \underline{0.0307} \\

\scenerule

% =========================================================
% Plush Toy (Trial 1)
% =========================================================
\cellcolor{gray!5}
& PhysTwin
& \underline{26.05} & \underline{0.9572} & \textbf{0.0389}
& \textbf{0.00613} & \textbf{0.0118}
& \underline{22.97} & \underline{0.9529} & \underline{0.0690}
& \underline{0.01337} & \underline{0.0256} \\

\cellcolor{gray!5}
& PGND
& 23.62 & 0.9507 & 0.0571
& 0.01628 & 0.0192
& 19.90 & 0.9451 & 0.0882
& 0.04337 & 0.0512 \\

\cellcolor{gray!5}
& GIC
& 18.63 & 0.9224 & 0.1225
& 0.04842 & 0.0844
& 17.98 & 0.9290 & 0.1153
& 0.05375 & 0.1141 \\

\rowcolor{blue!6}
\cellcolor{gray!5}
\multirow[c]{-4}{1.65cm}{%
\raggedright Plush Toy\\[-1pt](Trial 1)}
& \textbf{MorphTwin (Ours)}
& \textbf{28.00} & \textbf{0.9666} & \underline{0.0443}
& \underline{0.00615} & \underline{0.0124}
& \textbf{24.38} & \textbf{0.9601} & \textbf{0.0561}
& \textbf{0.01314} & \textbf{0.0251} \\

\scenerule

% =========================================================
% Plush Toy (Trial 2)
% =========================================================
\cellcolor{gray!5}
& PhysTwin
& \underline{28.41} & \underline{0.9696} & \underline{0.0266}
& \textbf{0.00515} & \textbf{0.0100}
& 22.89 & 0.9527 & \underline{0.0560}
& \textbf{0.01194} & 0.0361 \\

\cellcolor{gray!5}
& PGND
& 27.99 & 0.9694 & 0.0323
& 0.00636 & \underline{0.0110}
& 20.16 & 0.9423 & 0.0903
& 0.05357 & 0.0861 \\

\cellcolor{gray!5}
& GIC
& 26.85 & 0.9652 & 0.0551
& 0.00952 & 0.0167
& \underline{23.55} & \underline{0.9551} & 0.0670
& 0.01541 & \underline{0.0358} \\

\rowcolor{blue!6}
\cellcolor{gray!5}
\multirow[c]{-4}{1.65cm}{%
\raggedright Plush Toy\\[-1pt](Trial 2)}
& \textbf{MorphTwin (Ours)}
& \textbf{32.15} & \textbf{0.9825} & \textbf{0.0233}
& \underline{0.00552} & 0.0112
& \textbf{24.23} & \textbf{0.9580} & \textbf{0.0509}
& \underline{0.01283} & \textbf{0.0335} \\

\scenerule

% =========================================================
% Mousepad
% =========================================================
\cellcolor{gray!5}
& PhysTwin
& 20.70 & 0.9036 & 0.1599
& \underline{0.02745} & 0.0541
& 19.56 & 0.9134 & \underline{0.1449}
& \underline{0.01796} & \underline{0.0431} \\

\cellcolor{gray!5}
& PGND
& \underline{22.51} & \underline{0.9168} & \underline{0.1386}
& \textbf{0.01885} & \textbf{0.0298}
& \underline{21.43} & \underline{0.9143} & 0.1602
& 0.04839 & 0.0578 \\

\cellcolor{gray!5}
& GIC
& 20.96 & 0.9026 & 0.1617
& 0.03069 & 0.0729
& 17.04 & 0.8520 & 0.2052
& 0.02334 & 0.0895 \\

\rowcolor{blue!6}
\cellcolor{gray!5}
\multirow[c]{-4}{1.65cm}{\raggedright Mousepad}
& \textbf{MorphTwin (Ours)}
& \textbf{24.40} & \textbf{0.9531} & \textbf{0.1106}
& 0.02910 & \underline{0.0535}
& \textbf{23.40} & \textbf{0.9377} & \textbf{0.1185}
& \textbf{0.01645} & \textbf{0.0400} \\

\scenerule

% =========================================================
% Book
% =========================================================
\cellcolor{gray!5}
& PhysTwin
& \underline{23.92} & \underline{0.9098} & \underline{0.0729}
& \textbf{0.01008} & \textbf{0.0154}
& \underline{22.14} & \underline{0.9003} & \textbf{0.0785}
& \textbf{0.01641} & \textbf{0.0242} \\

\cellcolor{gray!5}
& PGND
& 23.28 & 0.9090 & 0.0761
& \underline{0.01404} & \underline{0.0163}
& 19.63 & 0.8951 & 0.1424
& 0.03358 & 0.0425 \\

\cellcolor{gray!5}
& GIC
& 18.51 & 0.8817 & 0.1802
& 0.04216 & 0.0674
& 19.30 & 0.8857 & 0.1792
& 0.02979 & 0.0551 \\

\rowcolor{blue!6}
\cellcolor{gray!5}
\multirow[c]{-4}{1.65cm}{\raggedright Book}
& \textbf{MorphTwin (Ours)}
& \textbf{25.93} & \textbf{0.9347} & \textbf{0.0711}
& \textbf{0.01008} & \textbf{0.0154}
& \textbf{23.02} & \textbf{0.9118} & \underline{0.0845}
& \underline{0.01642} & \underline{0.0243} \\

\scenerule

% =========================================================
% Water Bottle
% =========================================================
\cellcolor{gray!5}
& PhysTwin
& 32.41 & 0.9887 & \underline{0.0204}
& \textbf{0.00502} & \textbf{0.0078}
& \underline{28.61} & \underline{0.9826} & \underline{0.0324}
& \textbf{0.00644} & \underline{0.0120} \\

\cellcolor{gray!5}
& PGND
& \underline{33.85} & \underline{0.9901} & 0.0216
& 0.00604 & \underline{0.0079}
& 19.35 & 0.9599 & 0.1321
& 0.13740 & 0.1521 \\

\cellcolor{gray!5}
& GIC
& 28.80 & 0.9839 & 0.0354
& 0.01109 & 0.0172
& 21.98 & 0.9710 & 0.0652
& 0.02553 & 0.0404 \\

\rowcolor{blue!6}
\cellcolor{gray!5}
\multirow[c]{-4}{1.65cm}{\raggedright Water Bottle}
& \textbf{MorphTwin (Ours)}
& \textbf{35.02} & \textbf{0.9932} & \textbf{0.0128}
& \underline{0.00508} & 0.0080
& \textbf{30.07} & \textbf{0.9880} & \textbf{0.0209}
& \underline{0.00652} & \textbf{0.0117} \\

\scenerule

% =========================================================
% Sun Hat
% =========================================================
\cellcolor{gray!5}
& PhysTwin
& \underline{26.00} & \underline{0.8901} & \textbf{0.0925}
& \textbf{0.01084} & \textbf{0.0183}
& \underline{22.94} & \underline{0.8643} & \underline{0.1330}
& \textbf{0.01435} & \underline{0.0295} \\

\cellcolor{gray!5}
& PGND
& 23.84 & 0.8700 & 0.1413
& 0.02019 & 0.0288
& 19.92 & 0.8186 & 0.2296
& 0.03723 & 0.0702 \\

\cellcolor{gray!5}
& GIC
& 20.17 & 0.8200 & 0.2496
& 0.02921 & 0.0582
& 17.50 & 0.7601 & 0.3202
& 0.02727 & 0.0922 \\

\rowcolor{blue!6}
\cellcolor{gray!5}
\multirow[c]{-4}{1.65cm}{\raggedright Fabric Hat}
& \textbf{MorphTwin (Ours)}
& \textbf{28.63} & \textbf{0.9252} & \underline{0.0968}
& \underline{0.01146} & \underline{0.0194}
& \textbf{24.40} & \textbf{0.8889} & \textbf{0.1292}
& \underline{0.01487} & \textbf{0.0291} \\

\scenerule

% =========================================================
% Headphones
% =========================================================
\cellcolor{gray!5}
& PhysTwin
& \underline{28.94} & \underline{0.9721} & \underline{0.0245}
& \textbf{0.00588} & \textbf{0.0091}
& \underline{25.66} & \underline{0.9628} & \underline{0.0356}
& \textbf{0.00984} & \textbf{0.0132} \\

\cellcolor{gray!5}
& PGND
& 24.10 & 0.9545 & 0.0528
& 0.01631 & \underline{0.0184}
& 20.46 & 0.9346 & 0.0954
& 0.03057 & 0.0368 \\

\cellcolor{gray!5}
& GIC
& 23.02 & 0.9453 & 0.1036
& 0.04318 & 0.0934
& 18.98 & 0.9236 & 0.1607
& 0.06257 & 0.1461 \\

\rowcolor{blue!6}
\cellcolor{gray!5}
\multirow[c]{-4}{1.65cm}{\raggedright Headphones}
& \textbf{MorphTwin (Ours)}
& \textbf{32.05} & \textbf{0.9822} & \textbf{0.0180}
& \underline{0.00597} & \textbf{0.0091}
& \textbf{27.04} & \textbf{0.9689} & \textbf{0.0296}
& \underline{0.01000} & \underline{0.0135} \\

\scenerule

% =========================================================
% Smartphone
% =========================================================
\cellcolor{gray!5}
& PhysTwin
& \underline{32.16} & \underline{0.9870} & 0.0245
& \textbf{0.00524} & \textbf{0.0099}
& \underline{28.84} & \underline{0.9820} & \underline{0.0368}
& \underline{0.01530} & \underline{0.0217} \\

\cellcolor{gray!5}
& PGND
& 31.73 & 0.9864 & \underline{0.0230}
& 0.00719 & \underline{0.0106}
& 24.80 & 0.9753 & 0.0595
& 0.05315 & 0.0517 \\

\cellcolor{gray!5}
& GIC
& 31.54 & 0.9861 & 0.0351
& 0.02336 & 0.0447
& 20.81 & 0.9615 & 0.1132
& 0.06700 & 0.1465 \\

\rowcolor{blue!6}
\cellcolor{gray!5}
\multirow[c]{-4}{1.65cm}{\raggedright Smartphone}
& \textbf{MorphTwin (Ours)}
& \textbf{36.40} & \textbf{0.9924} & \textbf{0.0133}
& \underline{0.00587} & 0.0109
& \textbf{31.81} & \textbf{0.9873} & \textbf{0.0208}
& \textbf{0.01370} & \textbf{0.0197} \\

\bottomrule

\end{tabular}

\end{adjustbox}
\end{table*}

\endgroup

\begin{figure*}[htbp]
\centering
\includegraphics[width=0.8\linewidth]{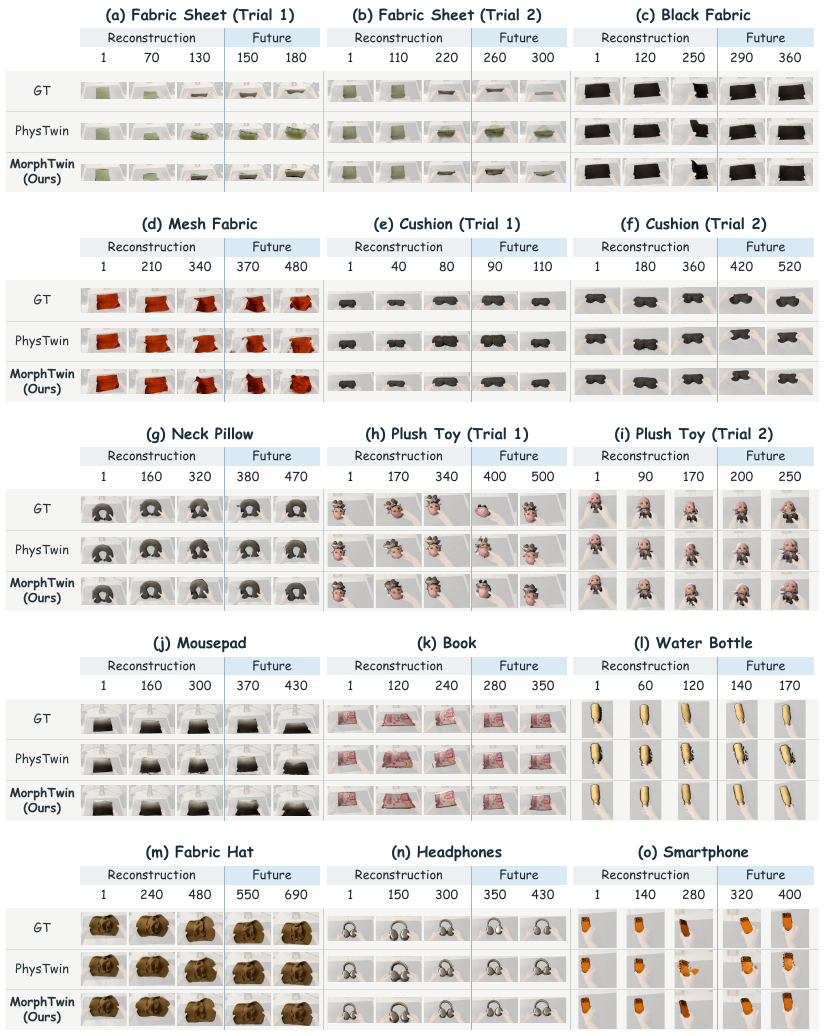}
\caption{
Reconstruction and future prediction on all 15 sequences of our benchmark under the All Views setting.
Each panel shows representative reconstruction and future frames for
GT, PhysTwin, and MorphTwin (Ours).
GT is displayed separately to facilitate direct comparison of object
motion and deformation across methods.
The corresponding per-scene quantitative results are reported in
Table~\ref{tab:morphtwin_per_scene_all_views}.
We strongly recommend viewing \textit{\underline{Supplementary Video-1}} for a clearer
comparison of motion and deformation over the full sequences.
}
\label{fig:morphtwin-per-scene}
\end{figure*}

\subsection{Material Editing and Reuse}
\label{app:material_reuse_fracture}

\paragraph{Cross-interaction reuse of physical models recovered from videos.}
We further test whether physical behavior recovered from one interaction
video can be reused for another interaction of the same object.
We evaluate
\textit{Fabric Sheet (Trial 2)}$\rightarrow$\textit{Fabric Sheet (Trial 1)}
and
\textit{Cushion (Trial 1)}$\rightarrow$\textit{Cushion (Trial 2)}.
For each pair, we reuse the physical model optimized from the source
video, including its spring network, carrier parameters, internal rest
lengths, masses, damping, and contact settings.
The source model is rigidly aligned to the target first frame by
vertical-axis rotation and translation, without scaling or non-rigid
alignment. Only the control-spring rest lengths are reset to the target
initial configuration.
The target hand trajectory is provided as the prescribed interaction
input, and the model is rolled out continuously over the full target
sequence without target specific physical optimization or a state reset
at the future boundary.
We compare against MorphTwin and PhysTwin models fitted locally on the
target recording.

\begin{table}[t]
\centering
\caption{
Cross-interaction reuse of physical models recovered from videos.
CD and Track are reported in millimeters (mm).
Transfer reuses the source-video physical model without
target specific physical fitting during rollout.
}
\label{tab:app_cross_interaction_transfer}

\footnotesize
\setlength{\tabcolsep}{2pt}
\renewcommand{\arraystretch}{1.08}

\begin{tabular}{@{}lcccc@{}}
\toprule
& \multicolumn{2}{c}{Modeling}
& \multicolumn{2}{c}{Future} \\
\cmidrule(lr){2-3}
\cmidrule(l){4-5}
Method
& CD $\downarrow$
& Track $\downarrow$
& CD $\downarrow$
& Track $\downarrow$ \\
\midrule

\multicolumn{5}{@{}l}{
\textit{Fabric Sheet (Trial 2)}
$\rightarrow$
\textit{Fabric Sheet (Trial 1)}} \\
MorphTwin transfer
& 10.05 & 20.62 & \textbf{25.19} & \textbf{54.19} \\
MorphTwin local
& \textbf{9.48} & \textbf{20.05} & 35.13 & 62.56 \\
PhysTwin local
& 10.02 & 23.06 & 31.82 & 62.88 \\
\midrule

\multicolumn{5}{@{}l}{
\textit{Cushion (Trial 1)}
$\rightarrow$
\textit{Cushion (Trial 2)}} \\
MorphTwin transfer
& 10.39 & 21.62 & 14.41 & 39.52 \\
MorphTwin local
& \textbf{5.14} & \textbf{10.14}
& \textbf{12.81} & \textbf{33.21} \\
PhysTwin local
& 8.65 & 14.41 & 16.06 & 39.45 \\
\bottomrule
\end{tabular}
\end{table}

The results are shown in Table~\ref{tab:app_cross_interaction_transfer}. For \textit{Fabric Sheet}, the transferred model remains close to the
locally fitted models on the observed interval and reduces future CD and
Track errors by $20.8\%$ and $13.8\%$, respectively, relative to local
PhysTwin.
For \textit{Cushion}, transfer has larger modeling errors, but its future
CD is $10.3\%$ lower than local PhysTwin and its future Track differs by
only $0.07$~mm, while the locally fitted MorphTwin remains more accurate.
The fabric case also exhibits run-to-run variation, so these numbers
should be interpreted as results from the evaluated transfer run.
These results demonstrate reuse of a fitted
physical asset across different interactions
of the same object.
The following controlled study examines a
distinct setting, transferring material responses
between different spring graphs.

\paragraph{Shared material responses across different graphs.}
Cross-graph material reuse aims to preserve
the source graph's measured bulk and shear
responses while adapting their spring realization
to the target graph.
We evaluate this principle using
\textit{apple}, \textit{E}, and \textit{rope},
with source graphs containing 2,048 particles
and target graphs containing 512, 1,024,
or 2,048 particles.
Three Poisson-ratio settings ($0.15$, $0.25$,
and $0.35$) yield 72 non-identity transfers
per method.

All graphs use unit total mass, so the reported
elastic scales are mass-normalized responses
rather than Pa-valued continuum moduli.
Let $\mathbf q_s^\star$ denote the logarithms
of the source graph's measured mass-normalized
bulk and shear responses.
The target graph seeks to match this response
pair, corresponding to a target-relative change
$\Delta\mathbf q_t^\star
=\mathbf q_s^\star-\mathbf q_t^0$
under the same normalization.

For finite-amplitude response matching,
each target graph uses a nonlinear response
table calibrated offline with virtual
small-strain equilibrium probes.
Spring parameters are obtained through
interpolation and stiffness rescaling
without fitting target motion.
This extends the graph-specific response
mapping in Sec.~\ref{sec:material_identification} beyond its local
linear approximation and directly evaluates
response preservation across geometries
and resolutions.

Table~\ref{tab:app_material_reuse} shows that the target graph’s calibration
reduces mean response error from $9.5173\%$ to $0.000503\%$ compared
with copying spring multipliers.
It also reduces energy error over 24 unseen strain directions from
$10.4833\%$ to $3.6203\%$.
These results support separating transferable mechanical response targets
from graph-dependent spring parameters: different geometries and
discretizations can realize the same bulk and shear responses through their
own calibration.

% Preamble:
% \usepackage{booktabs,adjustbox}
% \usepackage[table]{xcolor}
% \usepackage{amsmath}

\begin{table}[htbp]
\centering
\footnotesize
\setlength{\tabcolsep}{4pt}
\renewcommand{\arraystretch}{1.08}
\setlength{\heavyrulewidth}{0.45pt}
\setlength{\lightrulewidth}{0.30pt}

\caption{
Controlled response transfer across graphs on the
\textit{apple}, \textit{E}, and \textit{rope} geometries
from the Spring-Gaus dataset.
Three source material settings
($\nu=0.15,\,0.25,\,0.35$) on 2,048-particle graphs are transferred
to target graphs with 512, 1,024, or 2,048 particles.
Transfer targets are the measured mass-normalized bulk and shear response
values of the source graph under unit total mass.
Each strategy is evaluated on the same 72 non-identity conditions.
Response and energy errors are percentages; $\nu$ error is absolute.
Bold indicates the lower error within each scope.
}
\label{tab:app_material_reuse}

\begin{adjustbox}{max width=\linewidth}
\begin{tabular}{@{}llcccc@{}}
\toprule
Scope
& Transfer strategy
& Cases
& \shortstack{Response error\\(\%) $\downarrow$}
& $\nu$ error $\downarrow$
& \shortstack{Energy error\\(\%) $\downarrow$} \\
\midrule

\multirow{2}{*}{All transfers}
& Copy spring multipliers
& 72 & 9.5173 & 0.01435 & 10.4833 \\
& \cellcolor{blue!6}Target-graph calibration
& \cellcolor{blue!6}72
& \cellcolor{blue!6}\textbf{0.000503}
& \cellcolor{blue!6}$\boldsymbol{2.80\!\times\!10^{-7}}$
& \cellcolor{blue!6}\textbf{3.6203} \\

\addlinespace[3pt]
\multirow{2}{*}{Cross-object}
& Copy spring multipliers
& 54 & 9.3973 & 0.01560 & 10.3014 \\
& \cellcolor{blue!6}Target-graph calibration
& \cellcolor{blue!6}54
& \cellcolor{blue!6}\textbf{0.000482}
& \cellcolor{blue!6}$\boldsymbol{2.70\!\times\!10^{-7}}$
& \cellcolor{blue!6}\textbf{3.5529} \\

\addlinespace[3pt]
\multirow{2}{*}{Same object, new resolution}
& Copy spring multipliers
& 18 & 9.8773 & 0.01059 & 11.0291 \\
& \cellcolor{blue!6}Target-graph calibration
& \cellcolor{blue!6}18
& \cellcolor{blue!6}\textbf{0.000565}
& \cellcolor{blue!6}$\boldsymbol{3.12\!\times\!10^{-7}}$
& \cellcolor{blue!6}\textbf{3.8223} \\

\bottomrule
\end{tabular}
\end{adjustbox}
\end{table}

\begin{figure}[htbp]
\centering
\includegraphics[width=\linewidth]
{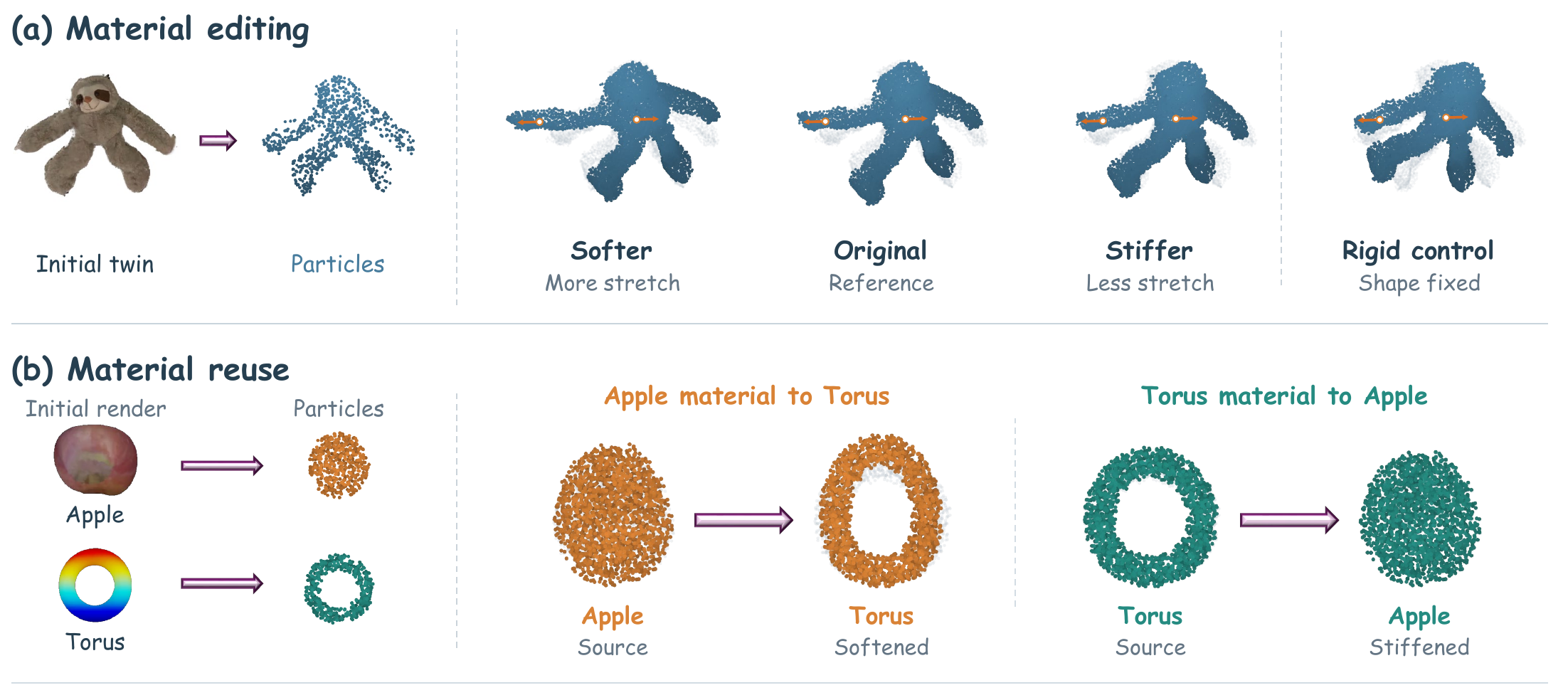}
\caption{
\textbf{Material editing and reuse across objects.}
Initial object renderings and their particle representations are shown
on the left.
\textit{(a) Material editing.}
Soft ($0.25\times$), original ($1\times$), and stiff ($4\times$)
settings produce progressively smaller deformations under identical
interaction forces, alongside a rigid-projection control.
The camera, scale, and observation time are shared across settings.
\textit{(b) Material reuse.}
Calibration for each graph realizes exchanged material response
targets on Apple and Torus, making Torus softer and Apple stiffer.
Particle colors indicate the source material settings.
Gray particles show the unloaded configuration.
Displacements in (b) are magnified $20\times$ for visualization.
}
\label{fig:app_material_particle}
\end{figure}

\paragraph{Material editing under fixed interaction forces.}
A reusable material representation should also support changing an
object's response without reconstructing its geometry.
Figure~\ref{fig:app_material_particle} (a) demonstrates this capability
under identical interaction forces.
Scaling spring stiffness from $0.25\times$ to $4\times$ reduces
the grip-span increase at $0.60$~s from $6.78$ to $1.47$~cm.
The original setting lies between these responses, while a separate
rigid-projection control preserves the grip span.
These results show how one reconstructed asset can support different
interaction behaviors through explicit material control, without
retraining geometry or appearance.

\paragraph{Exchanging materials while preserving geometry.}
We further exchange two material presets between the native
\textit{apple} and \textit{torus} graphs
(Figure~\ref{fig:app_material_particle} (b)).
Under the unit-total-mass convention described
above, the presets use mass-normalized elastic scales of $100$ and $400$ and
Poisson ratios of $0.15$ and $0.35$, respectively. These scale values represent
the stiffness-mass normalization used in this controlled experiment and
are not continuum Young's moduli in Pa.
Each object retains its geometry and connectivity.
For each donor preset, we measure its
mass-normalized bulk and shear responses
and use this response pair as the transfer target.
The receiving graph realizes these responses
through its own offline nonlinear response table,
using interpolation and stiffness rescaling
without fitting target motion.
Under fixed supports and loading within each object, exchanging
the material presets produces distinct particle deformations.
Across the two transfer directions, the target graph’s calibration reduces
bulk and shear responses error from $1.52\%$ and $4.90\%$
with copied multipliers to below $0.00013\%$.
This illustrates our central insight: material response targets
can be shared across objects while their spring realization
adapts to each object's graph.

\paragraph{Reusable Physical Twins for Robotic Interaction and Replay.}
We further demonstrate the reuse of reconstructed
physical assets in robotic interaction scenarios.
We import a Franka robot into NVIDIA Isaac Sim
and use predefined grasp trajectories to generate
robot motions through inverse kinematics.
Meanwhile, MorphTwin simulates the corresponding
object deformations using previously fitted physical
parameters, without refitting the assets.
The robot motions and simulated object deformations
are then synchronously replayed in Isaac Sim.
Figure~\ref{fig:isaac_sim_interaction} shows
interactive examples on three assets in our self-captured scenes.
See Supplementary Video~1 for the complete sequences.

\begin{figure}[htbp]
    \centering
    \includegraphics[width=\linewidth]{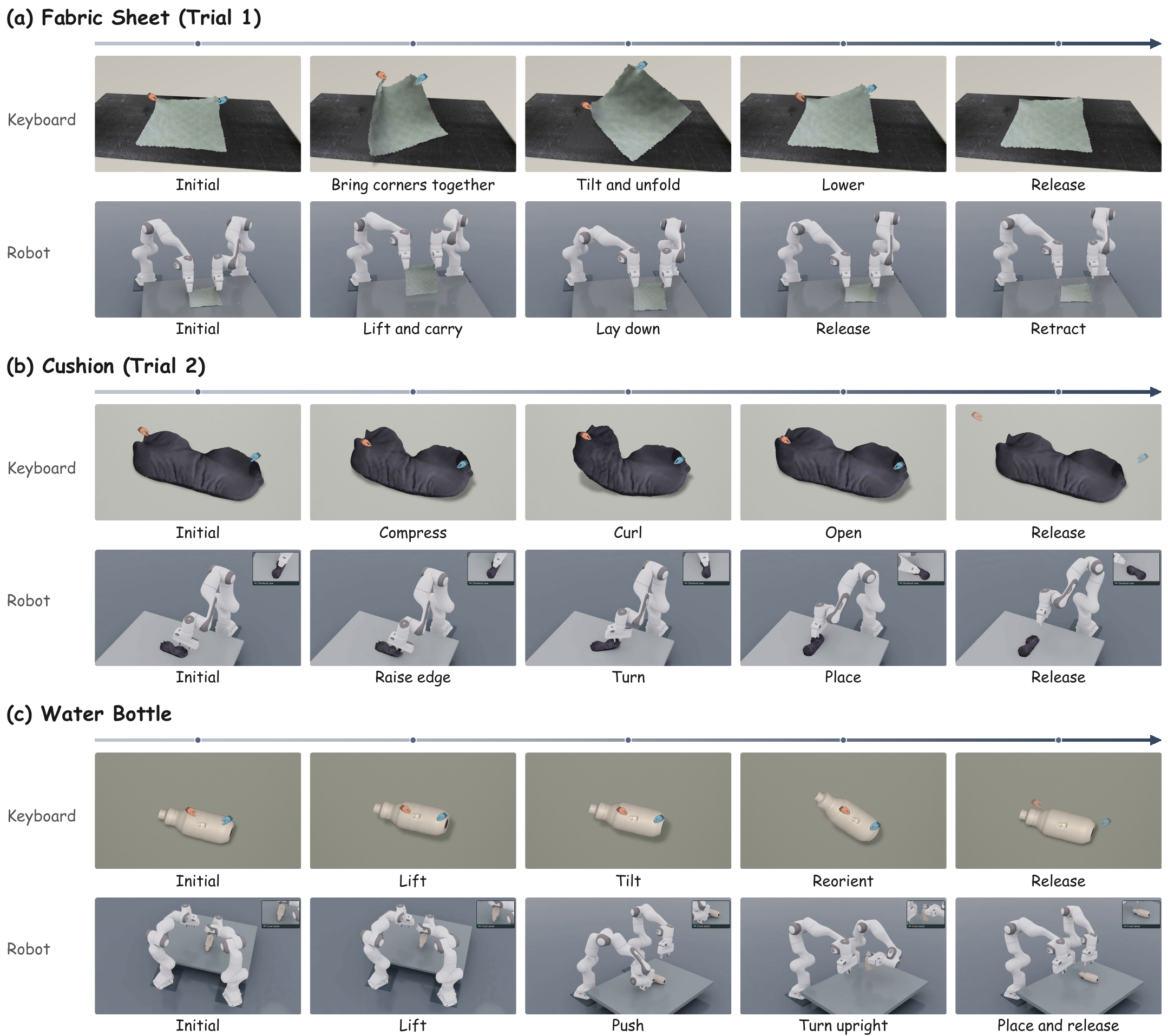}
    \caption{
    \textbf{Reusable physical assets for interactive control
    and robotic replay.}
    Previously reconstructed MorphTwin assets are reused
    under new interaction controls without refitting.
    We show (a) Fabric Sheet (Trial 1),
    (b) Cushion (Trial 2), and (c) Water Bottle.
    Top rows show keyboard-controlled simulations
    in MorphTwin.
    Bottom rows show synchronized replay in Isaac Sim,
    combining MorphTwin-simulated object deformations
    with Franka poses obtained via inverse kinematics
    from prescribed grasp trajectories.
    Arrows indicate temporal progression.
    }
    \label{fig:isaac_sim_interaction}
\end{figure}

\subsection{Fracture-aware Appearance}
\label{app:structural_damage}

Section~\ref{app:gaussian_binding} defines Gaussian deformation
using fixed reference neighborhoods of physical nodes.
While these neighborhoods provide local deformation support
during intact-graph simulation, they do not account for
connectivity changes caused by spring failure.
In MorphTwin, a failed spring removes its elastic contribution
and changes the surviving local connections available for
Gaussian appearance transport.
We therefore examine how appearance transport can adapt to
structural damage through a controlled fracture demonstration
and an ablation of connectivity-aware appearance binding.

\paragraph{Fracture with calibrated springs.}
Material calibration specifies the response of the intact graph.
Let $\mathbf Y_{\mathcal G}^{\mathrm{ref}}
=\{Y_e^{\mathrm{ref}}\}_{e\in\mathcal S}$
denote the graph-specific spring stiffnesses obtained from this
calibration.
We keep these stiffnesses and the reference rest lengths fixed
during fracture.
To model connection failure, we introduce a binary activity
variable $a_e\in\{0,1\}$ for each spring, where $1$ indicates
an intact spring and $0$ indicates a failed spring.
Let $\mathbf a=\{a_e\}_{e\in\mathcal S}$ collect all spring
activity states and $\mathbf x=\{\mathbf x_i\}_{i\in\mathcal P}$
denote the current positions of all simulation nodes.
The resulting elastic energy is
\begin{equation}
\begin{aligned}
U(\mathbf x,\mathbf a;\mathbf Y_{\mathcal G}^{\mathrm{ref}})
&=
\sum_{e=(i,j)\in\mathcal S}
a_e\,U_e(\mathbf x;Y_e^{\mathrm{ref}}),\\
U_e(\mathbf x;Y_e^{\mathrm{ref}})
&=
\frac{1}{2}Y_e^{\mathrm{ref}}L_e
\left(
\frac{\|\mathbf x_j-\mathbf x_i\|-L_e}{L_e}
\right)^2,
\end{aligned}
\label{eq:app_damage_energy}
\end{equation}
where $L_e$ is the reference rest length of spring $e$.
A failed connection contributes no elastic energy.
The stiffnesses of surviving springs remain fixed, so changes
in the effective structural response arise from the loss of
connections, without recalibrating the graph after each failure.

\begin{figure}[htbp]
\centering
\includegraphics[width=0.94\linewidth]
{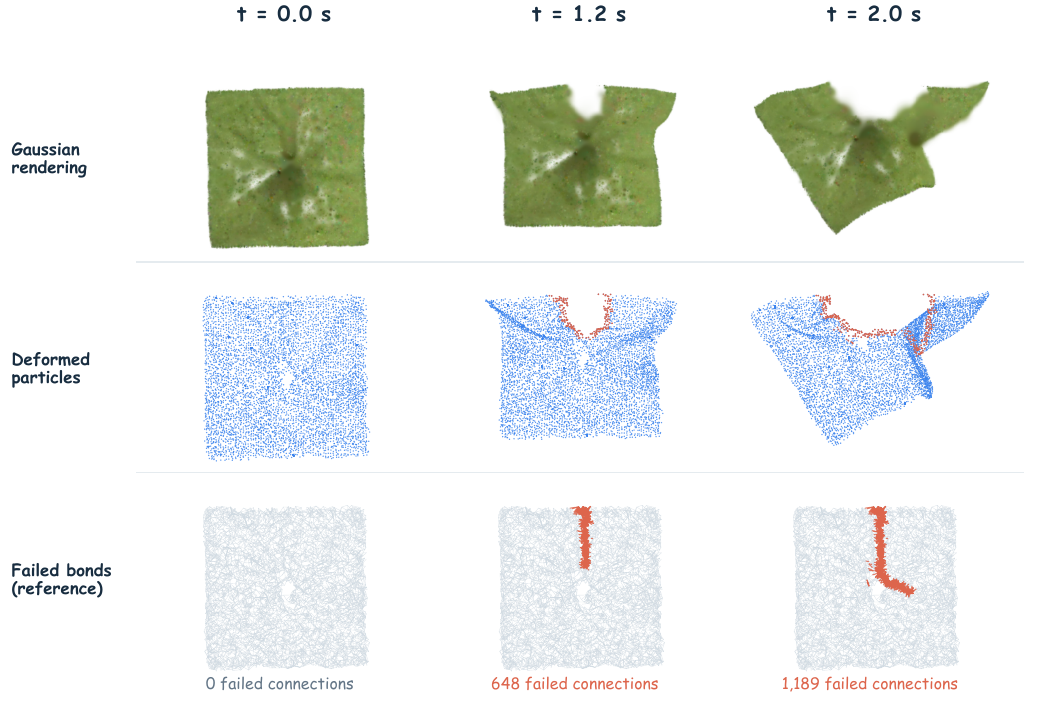}
\caption{
Progressive crack formation in the cloth twin under two opposing
grasps.
Columns show successive interaction stages.
Top: Gaussian renderings show the evolving cloth appearance.
Middle: deformed particles, with failed-connection endpoints
highlighted in orange.
Bottom: failed connections in reference coordinates reveal the
spatial progression of damage independently of deformation.
The main component remains connected as the crack opens,
motivating appearance bindings that follow surviving local
connections before complete separation.
}
\label{fig:app_fracture_progression}
\end{figure}

We demonstrate this behavior on the cloth twin using two opposing
grasps of radius 36~mm, each moving 12~cm along a half-cosine
trajectory.
The controller force is capped at 40~N per grasp, with unit total
mass and gravity disabled.
Connections fail irreversibly when their tensile strain exceeds
the prescribed threshold of $0.35$.
Figure~\ref{fig:app_fracture_progression} shows the resulting
Gaussian appearance, deformed particles, and failed connections.
At 1.2~s and 2.0~s, respectively, 648 and 1,189 connections
have failed, while the main connected component retains its
5,881 nodes.
Local cracks therefore develop before the object separates into
disconnected components, motivating appearance updates that
follow local changes in connectivity.

\paragraph{Local connectivity for appearance transport.}
We isolate the effect of local connectivity by comparing three binding strategies: (i) fixed $K_{\mathrm{bind}}=16$ binding, which assigns each appearance tracer to its 16 nearest physical nodes in the reference configuration; (ii) connected-component filtering, which retains only candidate nodes belonging to the same connected component as the original binding neighborhood; and (iii) surviving-local-support filtering, which further restricts candidates to those supported by surviving local connections.
The evaluation uses grids with a partial slit and a remaining
connection bridge.
We evaluate nine combinations of three grid resolutions
($16\times16$, $24\times24$, and $32\times32$)
and three opening widths (5, 10, and 20~mm).
Each condition uses 4,096 appearance tracers and the same prescribed
displacement field, with analytic tracer positions as the reference.
Initial binding candidates, weights, regularization, and deformation
limits are shared across methods.

Table~\ref{tab:app_binding_ablation} shows that restricting bindings
to surviving local supports reduces mean tracer error from
$0.313$ to $0.049$~mm and P95 error from $1.957$ to $0.305$~mm.
Component-only restriction gives the same results as fixed binding
because the grid remains globally connected.
These results show that local connectivity helps avoid interpolation
across an opening crack before complete separation.
The explicit graph thus serves both as the realization of calibrated
material responses and as the structural support for appearance
updates as connections fail.

\begin{table}[t]
\centering
\small
\setlength{\tabcolsep}{2.5pt}
\renewcommand{\arraystretch}{1.08}
\setlength{\heavyrulewidth}{0.45pt}
\setlength{\lightrulewidth}{0.30pt}

\caption{
Controlled binding diagnostic over nine conditions.
Errors are in millimeters relative to the analytic reference
and are equally averaged across conditions;
P95 is computed per condition.
}
\label{tab:app_binding_ablation}

\begin{tabular}{@{}lccc@{}}
\toprule
Binding support
& Mean $\downarrow$
& P95 $\downarrow$
& Near-slit $\downarrow$ \\
\midrule

Fixed $K_{\mathrm{bind}}=16$
& 0.313 & 1.957 & 1.940 \\

Component only
& 0.313 & 1.957 & 1.940 \\

\rowcolor{blue!6}
Surviving local support
& \textbf{0.049}
& \textbf{0.305}
& $<10^{-9}$ \\

\bottomrule
\end{tabular}
\end{table}

\begin{figure}[htbp]
\centering
\includegraphics[width=0.98\linewidth]{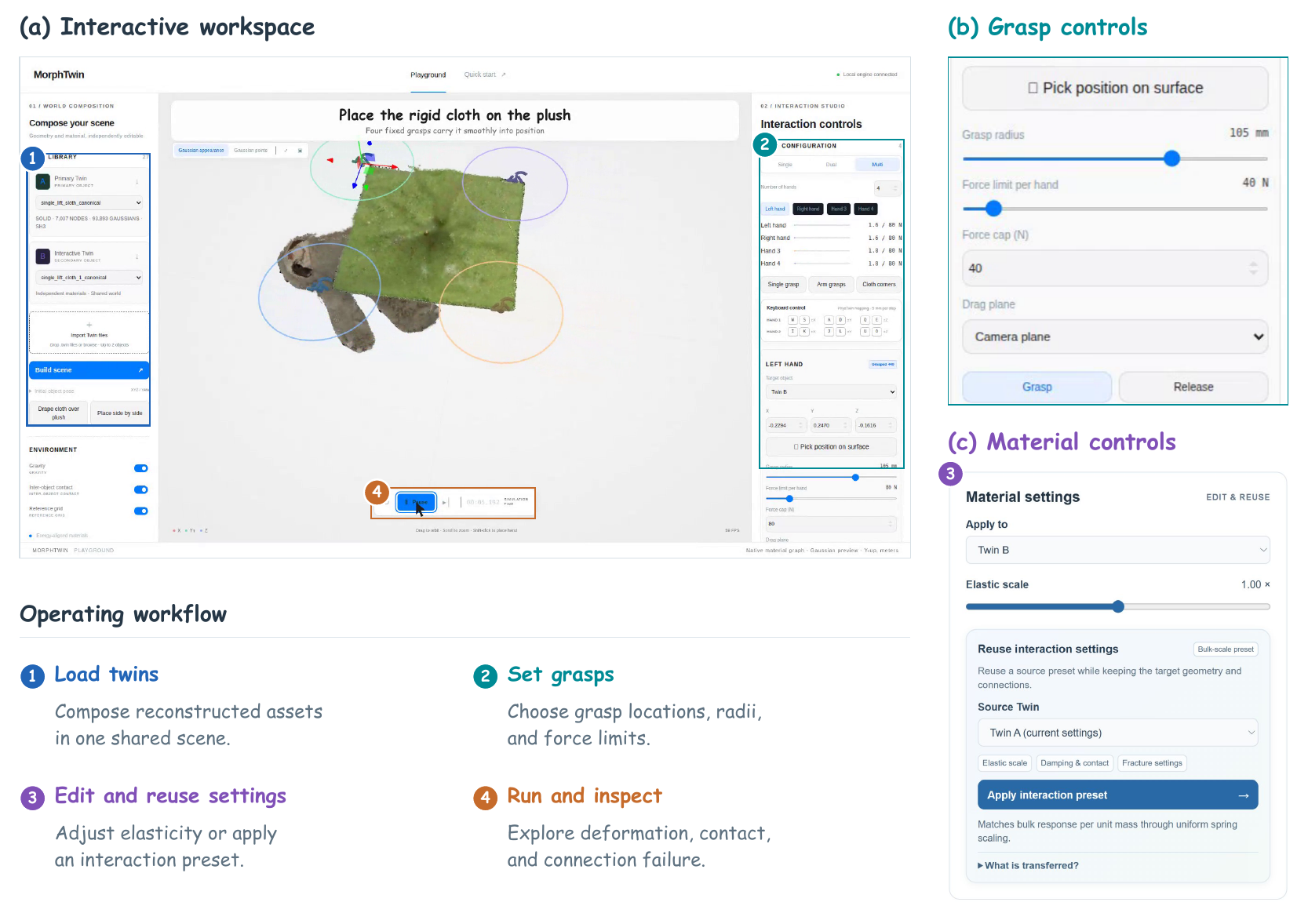}
\caption{\textbf{MorphTwin's interactive interface for composing and
manipulating reconstructed physical twins.}
\textit{Part (a):} Shared workspace; numbered markers link controls to
the operating workflow.
\textit{Part (b):} Grasp controls for selecting positions, adjusting
radii and force limits, and grasping or releasing objects.
\textit{Part (c):} Elastic editing and interaction-preset reuse.}
\label{fig:app_interactive_ui}
\end{figure}

\section{Interactive interface}
\label{app:ui}

Figure~\ref{fig:app_interactive_ui} illustrates the interaction workflow. 
Users load reconstructed twins into
a shared scene and manipulate them through single, dual, or multiple
grasps, with adjustable locations, radii, and force limits. They can
adjust elastic scale, configure rigid and fracture behavior, and reuse
interaction presets while retaining the target's reference geometry and
connectivity. Running or pausing the explicit simulation lets users inspect
deformation, contact, and connection failure under new manipulations.

MorphTwin's response-preserving material reuse operates in the shared
volumetric and shear response space and uses the target graph’s energy calibration,
as evaluated in our material-transfer experiments. The interface's
preset-import shortcut instead matches a bulk-response scale and copies
interaction settings; it is distinct from this calibrated transfer.

\end{document}